# Meta-Analysis with Untrusted Data

Shiva Kaul [†]  
skkaul@cs.cmu.edu

Geoffrey J. Gordon [*]  
ggordon@cs.cmu.edu

**Abstract**

Meta-analysis a crucial tool for answering scientific questions. It is usually conducted on a relatively small amount of "trusted" data, ideally from well-conducted, randomized, controlled trials. Each trial $i$ has features $X_i$ (describing its intervention and population), a true (unobserved) causal effect $U_i$, and an observed effect $Y_i \sim N(U_i, V_i)$, where $V_i$ is within-trial variance due to a limited number of participants. Given new features $x$, the task is to predict an interval which contains the true effect $u$ with high probability. Traditional algorithms ignore the features (which induce heterogeneous effects) because the amount of trusted data is too small to train a predictor of $u$ given $x$. They make poor predictions when there is a lot of heterogeneity. In this paper, we consider using untrusted data drawn from large observational databases, related scientific literature and practical experience. The (seemingly audacious) goal is to maintain rigorous prediction of causal effects, despite potential confounding and errors in the untrusted data.

We achieve this goal with two new algorithms based on conformal prediction. This methodology fundamentally produces rigorous prediction intervals, even when incorporating untrusted data, but it does not address the problem of handling noise: we seek intervals for the new $u$, but observe only $Y$ and $V$. The first algorithm is appropriate when the trials are very large (i.e. $V_i \approx 0$), so meta-analysis is close to noise-free prediction. This algorithm uses an underlying noise-free conformal predictor (which takes $U$ and predicts an interval for $u$) and determines its worst-case predictions over all plausible $\widehat{U}$ near $Y$. The second algorithm is designed for the more practical, challenging case of smaller trials. Given a new trial's $x$ and $v$, we can conformally predict an interval for $y \sim N(u, v)$; the problem is, we don't know $v$. At any possible $v$, we show the interval for $y$ can have $O(\sqrt{v})$ width "shaved off" to produce an interval for $u$. Then, we take the worst-case $v$: larger $v$ leads to a larger interval for $y$, but more of it can be shaved. In order for both of these algorithms to work, the underlying conformal prediction algorithm must be analytically tractable and capable of incorporating prior information (the untrusted data). For these purposes, we use fully-conformal kernel ridge regression. Under a novel condition called *idiocentricity* (which can be ensured by taking the ridge parameter $\lambda$ reasonably large), we show that fully-conformal KRR is simple and efficient.

On multiple biomedical datasets, we show that conformal meta-analysis resolves the challenge of heterogeneity, delivering much tighter intervals than traditional algorithms. Furthermore, it improves the rigor of such predictions, since traditional algorithms are based on large-sample approximations with no finite-sample guarantees. It demonstrates qualitative improvements in a case study using real data from a published meta-analysis on amiodarone. Overall, this paper charts a radically new course for meta-analysis and evidence-based medicine, where heterogeneity and untrusted data are embraced to deliver more nuanced and precise predictions.

---

[†]Computer Science Department, Carnegie Mellon University, Pittsburgh PA 15213  
[*]Machine Learning Department, Carnegie Mellon University, Pittsburgh PA 15213

# 1 Introduction

A systematic review of a scientific question formally collects relevant, reliable evidence and answers the question as precisely as the evidence allows. Roughly 30,000 systematic reviews are currently published every year, either as standalone scientific papers or as part of clinical practice guidelines, by thousands of academic, professional, and regulatory organizations [Hoffmann et al., 2021]. Systematic reviews of the following questions each have over 2000 citations:

- Do anti-TNF antibodies for rheumatoid arthritis increase the risk of serious infections and malignancies? [Bongartz et al., 2006]
- Does bariatric surgery improve or resolve the clinical and laboratory manifestations of type 2 diabetes mellitus? [Buchwald et al., 2009]
- Can oral and topical pharmacotherapies treat neuropathic pain? [Finnerup et al., 2015]
- Does exercise-based cardiac rehabilitation improve cardiovascular mortality in patients with coronary heart disease? [Anderson et al., 2016]
- How effective is nicotine replacement therapy (gum, patches, spray, etc.) in achieving abstinence from smoking? [Hartmann-Boyce et al., 2018]

Systematic reviews adhere to highly-scrutinized methodology [Higgins et al., 2019, Schünemann et al., 2013, Page et al., 2021] and are widely considered to be the pinnacle of empirical evidence [Guyatt et al., 1995, Murad et al., 2016]. They have a decisively influential role in healthcare and related fields, especially in contentious situations where different parties disagree or have competing interests. This is because systematic reviews are designed to be rigorous and unbiased, in a broad sense [Sackett, 1979]: they should yield reliably correct answers, unblemished by personal opinions, conflicts of interest, unproven assumptions, and/or confounding of causation by correlation.

*Meta-analysis* is the statistical core of most systematic reviews. A key goal of meta-analysis[1] is to learn, from the collected evidence, a predictor $C$ of causal effect: given features $x$ of a treatment (encompassing its population, intervention, comparison and outcome measure), the predicted interval $C(x) \subseteq \mathbb{R}$ should, with high probability, contain the true effect $u$ of the treatment. As described here, meta-analysis models heterogeneity in the treatment and (in turn) its effect. For example, changing the age of patients or the dosage of a drug corresponds to a change in $x$, which would lead to a possibly different prediction $C(x)$ of a different $u$. Unfortunately, prevalent meta-analysis algorithms do not model heterogeneity, treating its consequences as inexplicable random noise in $u$. This is because the complexity of meta-analysis is profoundly constrained by the stringent expectations placed upon systematic reviews: only a limited fraction of "trusted" evidence is allowed in meta-analysis, which limits the kind of analysis that can actually be conducted.

Specifically, to avoid the confounding biases of observational studies, meta-analysis is (ideally) based solely upon well-conducted, randomized, controlled trials. These allow causal questions (e.g. "what is the effect of administering this drug?") to be reliably answered. On average, about 10-20 RCTs are included in the meta-analysis of a systematic review [Hoffmann et al., 2021]. Note that modern meta-analyses, especially network meta-analyses, can be much larger; for example, recent meta-analyses of glaucoma treatments, antipsychotics and antidepressants included 114, 402 and 522 trials, respectively [Li et al., 2016, Huhn et al., 2019, Cipriani et al., 2018]. Regardless, these data represent a tiny fraction of the accumulated human experience with the empirical phenomena of interest. Observational studies drawn from large databases now routinely involve millions of patients [Hripcsak et al., 2016, All of Us Research Program Investigators, 2019]. The personal experiences and opinions of practitioners, though less formal, are even more expansive.

---

[1]Meta-analysis also involves tasks such as estimating parameters with confidence intervals; see Section 2.2.

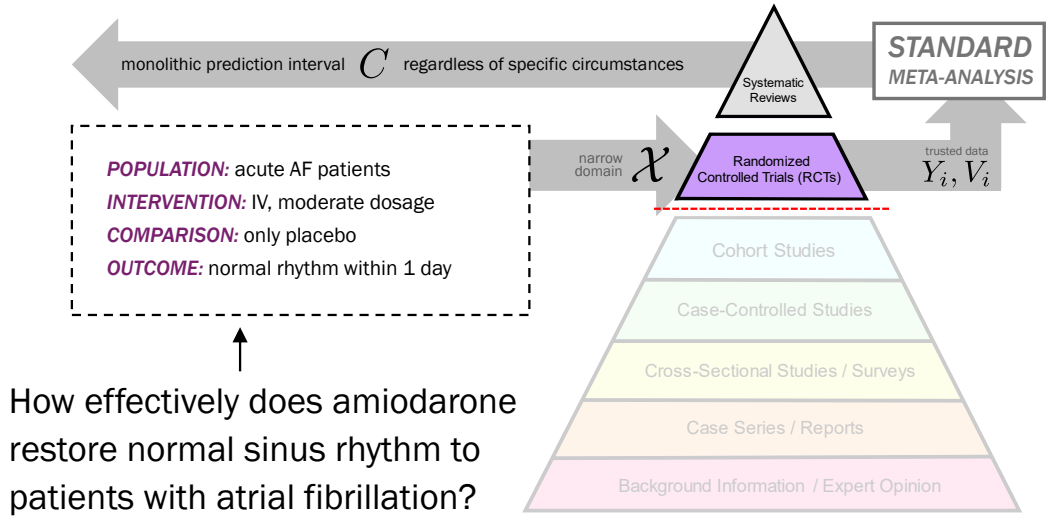

Figure 1: How meta-analysis presently answers scientific questions. The fundamental issues are that (1) untrusted data, constituting the majority of the evidence hierarchy [Guyatt et al., 1995], are not utilized, and (2) heterogeneity in the treatments (and their effects) is not explained. To avoid heterogeneity, meta-analyses are limited to relatively narrow domains $\mathcal{X}$. Only trusted data relevant to $\mathcal{X}$ are included. This small amount of data makes meta-analysis primitive and inflexible: it returns a single prediction interval $C$ which does not change based on specific treatment circumstances, such as the age of the patients or the dose of the drug.

Due to the paucity of trusted data, there is little hope in learning the complex relationship between $x$ and $u$. Present algorithms ignore $x$, so their (monolithic) prediction interval $C(x)$ must be wide enough to accommodate nearly all $x$. This inflexibility makes it difficult to establish good scientific evidence in fields with heterogeneous treatments. For example, in exercise science, the treatment effect may strongly depend on a large number of variables such as frequency, duration, equipment, technique, age, and diet [Rippetoe, 2017, Ferreira et al., 2010]. In most psychological research, the variation in $u$ is attributable primarily to between-study heterogeneity rather than within-study sampling variance [Stanley et al., 2018]. Some researchers believe that modeling heterogeneity is essential to progress in behavorial science, heralding such statistical advancement as a "heterogeneity revolution" [Bryan et al., 2021]. They echo the hope of using covariates expressed by DerSimonian and Laird [1986] in their seminal paper on random-effects meta-analysis.

## 1.1 Conformal Prediction

This paper demonstrates that untrusted data — with all its possible confounding, biases, and even outright errors — can be incorporated into meta-analysis while remaining rigorous and unbiased. In fact, this paper offers stronger, provable guarantees while weakening the assumptions traditionally employed in meta-analysis. The solution is based upon *conformal prediction* [Vovk et al., 2005, Shafer and Vovk, 2008, Lei and Wasserman, 2014]. The core idea of conformal prediction is that the residual of a future trial probably isn't extremely large compared to residuals of past trials.

Given $x$, the conformal prediction interval $C(x)$ simply excludes those $u$ which would lead to an extreme residual for a (hypothetical) future trial $(x, u)$. Though only the trusted data are used for these residual comparisons, the untrusted data can be incorporated into the underlying learning algorithm producing the residuals. If they help produce an accurate predictor — that is, one with small residuals — then the resulting interval $C(x)$ is tight, because small changes to $u$ dramatically increase the rank of the residual. If the untrusted data don't align with the trusted data, then $C(x)$ widens, but it remains correct, as it contains $u$ with the required probability.

While conformal prediction aptly manages the inclusion of untrusted data, there are two unresolved challenges when applying it to meta-analysis. The first challenge is noise: though we aim to predict true effects $u$, the empirically observed effects $y$ are blurred by limited trial sizes. Mathematically, this is modeled as Gaussian noise $y \sim N(u, v)$, where $v$ may depend on $x$ and $u$. This noise is curiously challenging to manage, since small (high noise) studies can differ fundamentally from large (low noise) studies. This reflects difficulties in clinical practice, where large-scale trials routinely fail to confirm the results of smaller ones [Ioannidis, 2005, Komajda et al., 2010, Manson et al., 2019]. This problem is serious enough that meta-analyses attempt to detect it using funnel plots [Light and Pillemer, 1984], whose validity is controversial [Lau et al., 2006].

The second challenge arises from limited sample size[2] of $n \leq 500$. When conformal prediction is applied in practice, the residuals are usually computed on a held-out sample of data, to avoid invalid comparisons between the $n$ training residuals and the 'fresh' test residual for the $(n+1)$'th trial. Such split conformal prediction is not feasible when $n$ is small, since we cannot afford to partition the data. A less common, but more efficient, approach is full conformal prediction. This turns the test residual into a training residual, making the comparisons valid again, by running the learning algorithm on all $n + 1$ samples — for all hypothetically possible values of $u$. This may pose a severe computational burden, and complicates efforts to handle noise.

## 1.2 Our Contributions

This paper resolves the aforementioned challenges of applying conformal prediction, giving rise to *conformal meta-analysis*. It culminates in two meta-analysis algorithms, which both consist of the following layers: (1) a representation of untrusted data as a prior probability distribution over plausible relationships between $x$ and $u$, (2) a learning algorithm which takes the prior and the training data, and returns an updated posterior distribution, (3) an efficient and simple implementation of full conformal prediction, operating upon residuals produced by the learning algorithm, and (4) a strategy for handling noise, exploiting the simplicity of the underlying conformal prediction.

Layers (1) and (2) are familiar: the untrusted data are represented as a Gaussian process, defined by a mean function $\mu(x)$ and a kernel function $\kappa(x, x')$. The learning algorithm is Gaussian process regression, also known as Bayesian inference with a conjugate Gaussian process prior and a normal likelihood, or kernel ridge regression (KRR). This algorithm is capable of arbitrarily-powerful nonlinear regression and can incorporate essentially any source of untrusted data. It has a hyperparameter $\lambda$ corresponding to either the variance of the likelihood or the ridge penalty. Thus, layer (3) amounts to fully-conformal KRR. Burnaev and Nazarov [2016] already derived fully-conformal KRR for general $\lambda$. Though their algorithm is computationally efficient, it returns a general prediction set (a union of disjoint intervals and isolated singletons) which isn't amenable to analytic reasoning. We show that, for practically sensible settings of $\lambda$, these complexities can be eliminated. This is because sufficiently high $\lambda$ makes KRR *idiocentric*: as $u$ varies, the residual for $(x, u)$ changes more than the other residuals. Under this condition, and some mild approximation, fully-conformal KRR can be simplified to computing quantiles in two lists.

---

[2]Modeling heterogeneity allows more expansive questions to be reviewed, generally leading to larger $n$.

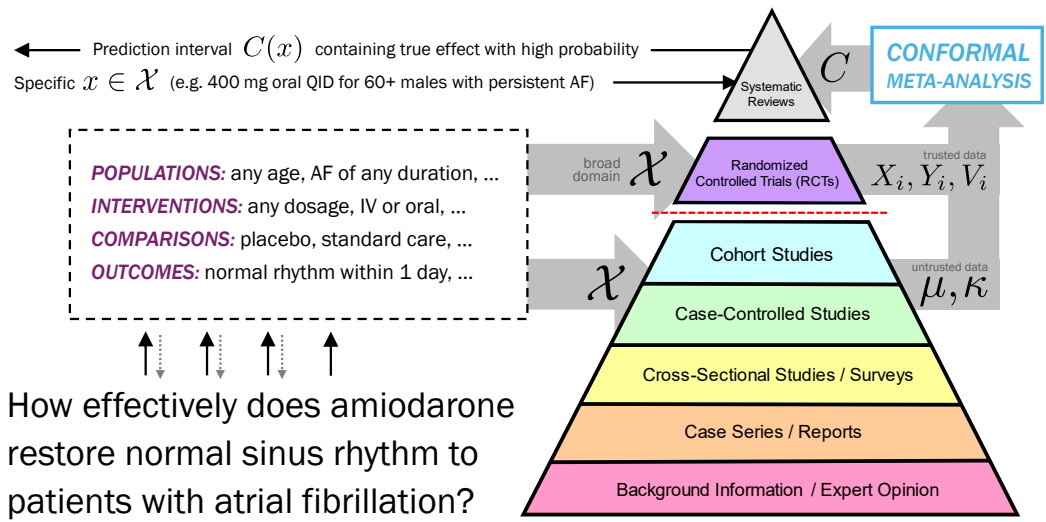

Figure 2: This paper changes how meta-analysis answers scientific questions. First, a relatively broad domain $\mathcal{X}$ for the meta-analysis is determined, possibly through further interaction with the user. This allows more expansive questions which include more data. Next, both trusted and untrusted data relevant to $\mathcal{X}$ are retrieved. Conformal meta-analysis takes these and produces not just a single interval, but a predictive model $C$. Given specific treatment circumstances $x$, the model predicts an interval $C(x)$ which, under standard assumptions, contains the true effect with high probability.

If we had observed true effects $U$ rather than noisy $Y$, then just running fully-conformal KRR would yield a satisfactory interval. We don't know $U$, but we do know the variances $V$, so we know that, outside a rare event of probability $\delta$, $U$ lies in some region $\mathcal{U}$ around $Y$. Thus, the statistical problem of covering $u$ reduces to a purely computational problem: bound the intervals generated by all plausible $\widehat{U}$ within $\mathcal{U}$. This can be formulated as an integer quadratic program, whose notional computational content is somewhat daunting: an infinite-dimensional regression is run an infinite number of times, for each possible $u$ in the interval, and then the range of such intervals is bounded over infinitely many $\widehat{U}$. Fortunately, due to the structure of fully-conformal KRR, the program can be bounded by a simple matrix norm computation. This bound is tight for $V \approx 0$. However, due to this algorithm's somewhat blunt and pessimistic statistical analysis, it suffers when meta-analysis departs from regression (i.e. there are trials having $V_i \gg 0$).

The second algorithm is more robust to trials of limited size. Rather than reasoning about the worst possible training input $U$ when conformally predicting $u$ as $C(x)$, the second algorithm adopts the complementary strategy: it reasons about the worst possible test input $v$ when conformally predicting $y$ as $C(x, v)$. In the latter approach, both $V$ and $v$ are presented to fully-conformal KRR, so its residuals can incorporate noise-correcting terms. More noise tends to produce larger residuals, so we naturally subtract more from residuals with higher $V_i$. This approach effectively reduces the importance of small trials, whereas the previous algorithm had to consider a large range for their possible true effects. A parameter $\eta > 0$ controls the extent of noise correction. $\eta = 1$ subtracts the full expectation of the noise. Smaller $\eta$ is less corrective, but has an advantage: the interval $C(x, v)$ is only $\sqrt{\eta v}$ wider than $C(x, 0)$. $\eta$ therefore controls the aggressiveness of the following strategy: since we don't know $v$, just return $C(x, 0)$, "shaving" the $\sqrt{\eta v}$ growth around it. Since shaving might inadvertently cut off some $u$, fully-conformal KRR must be run with a higher underlying confidence level, which requires larger $n$. Thus, conformal meta-analysis is feasible with a small number of large trials, but not a small number of small trials.

Our experiments have two goals: (1) to quantify how much conformal meta-analysis could improve predictions when used, as intended, with large amounts of untrusted data, and (2) to more qualitatively assess, before such data are available, how it would impact the experience of producing and consuming systematic reviews. At a high level, we find that conformal meta-analysis could improve how the medical community interacts with evidence.

## 2 Preliminaries

These are the predictive goals of meta-analysis.

**Predicting Effects.** *Let $\mathbb{P}$ be a distribution over features $x \in \mathcal{X}$, true effects $u \in \mathbb{R}$ and variances $v > 0$. For $i \in \{1, \ldots, n\}$, let $(X_i, U_i, V_i)$ be exchangeable samples from $\mathbb{P}$. Let $Y_i = U_i + \mathcal{E}_i$, where $\mathcal{E}_i \sim N(0, V_i)$ are independent. Let $\mu : \mathcal{X} \to \mathbb{R}$ and $\kappa : \mathcal{X} \times \mathcal{X} \mapsto \mathbb{R}$ be fixed (with respect to $\mathbb{P}$) mean and positive-definite kernel functions, respectively. From $\mu$, $\kappa$, $X$, $Y$, and $V$, produce a prediction band $C$ such that $\mathbb{P}(u \in C(x)) \geq 1 - \alpha$ for a desired confidence level $\alpha \in (0, 1)$.*

**Predicting Trials.** *Same as above, except $C$ also takes $v$, and should satisfy $\mathbb{P}(y \in C(x, v)) \geq 1 - \alpha$, where $y = u + \epsilon$ for independent $\epsilon \sim N(0, v)$.*

The first task is more practically useful and technically involved. However, since $u$ is not observable, but $y$ is, the second task is more easily verifiable. It is not immediately clear which task is more challenging, in the sense of needing wider intervals. On one hand, $y$ has inherently more variance than $u$. On the other, the prediction of $u$ is made without knowing $v$, which might otherwise distinguish between small and large trials having characteristically different $u$. The rest of this section is a review, thoroughly describing the origin and purpose of these tasks. It can be skipped by readers interested solely in the novel technical developments of this paper.

### 2.1 Outcomes and Effects

Let $x \in \mathcal{X}$ be features describing a treatment. This consists of the prospectively-set criteria of its population, intervention, comparison, and measure of outcome, commonly abbreviated as PICO [Richardson et al., 1995]. For example, $x$ may include the duration of an exercise program and the minimum age of its participants. It may also include auxiliary information that was collected passively and retrospectively, though (as described in the next section) this may complicate the interpretation of the meta-analysis. $x$ does not have to be numerical; it can be, for example, a published document describing a clinical trial. The number of participants in such a trial should not be intentionally encoded in $x$, since a treatment should be applicable to any number of people. However, avoiding implicit, unintentional correlations between trial design and trial size may be difficult or impossible. Let $\xi$ encode factors which influence the treatment, but are neither controlled nor observed. For example, the effect of an exercise program may surreptitiously depend on the altitude of the training facility or the jobs of the participants.

In the Neyman-Rubin framework of potential outcomes [Neyman, 1923, Rubin, 1974], for a single participant denoted by $\rho$, $\rho(1) \in \mathbb{R}$ is the outcome when assigned the treatment, and $\rho(0) \in \mathbb{R}$ is the outcome when assigned the comparison. Each outcome may be a final measurement (such as the amount of strength gained after training), or its change from a baseline measurement, or the logarithm of the ratio of final to baseline. The difference $\rho(1) - \rho(0)$ is the individual effect of the treatment. The potential outcomes framework is challenging because we cannot observe both terms in $\rho(1) - \rho(0)$, since each participant is assigned to either the treatment or the comparison. The conditional average treatment effect (CATE), denoted by $u$, quantifies the expected difference between the treatment and comparison for a new participant:

$$u(x, \xi) = \mathbb{E}_\rho \left( \rho(1) - \rho(0) \mid x, \xi \right) \qquad (1)$$

## 2.2 Different Goals of Meta-Analysis

The CATE is the predictive target of meta-analysis. With high probability (typically 95%, with $\alpha = 0.05$), the CATE should lie within the predicted interval:

$$\mathbb{P}_{C,x,\xi} \left( u(x, \xi) \in C(x) \right) \geq 1 - \alpha \qquad (2)$$

Rather than predicting relatively specific, tangible effects, meta-analysis often focuses on estimating more abstract, harder-to-verify quantities. Meta-analyses usually report a confidence interval $\mathrm{CI} \subset \mathbb{R}$ which, with high probability, should contain the average treatment effect (ATE, also known as the summary effect or grand mean):

$$\mathbb{P}_{\mathrm{CI}} \left( \mathrm{ATE} \in \mathrm{CI} \right) \geq 1 - \alpha \quad \text{where} \quad \mathrm{ATE} = \mathbb{E}_{x,\xi}\ u(x, \xi)$$

Whereas the confidence interval merely needs to capture the ATE, the prediction interval must capture most of the dispersion around it. (Formally, a prediction interval covers a random variable, and its coverage probability must also account for the randomness of that variable, whereas a confidence interval covers a fixed value). In the presence of significant heterogeneity, the confidence interval is much tighter than the prediction interval, and has little chance of capturing the effect of a future treatment. Due to this potentially unintuitive behavior, and the possibility of instilling overconfidence in evidence about the treatment, many prominent researchers encourage systematic reviews to report prediction intervals [IntHout et al., 2016, Riley et al., 2011, Borenstein, 2024]. According to some researchers, the relative ease of corroborating (or refuting) predictions makes them essential for scientific rigor and reproducibility [Billheimer, 2019].

These problems are exacerbated by the introduction of features ($x$) and larger numbers of trials ($n$), as proposed in this paper. Since confidence intervals are tighter than prediction intervals, it may be technically tempting to use untrusted priors to analogously tighten intervals for ATE. However, when considering many trials with substantially different features, ATE becomes a useless quantity [Simonsohn et al., 2022, Subramanian et al., 2018, Gould, 2010]. It is arguably misleading to use features within a statistical analysis but to simultaneously obfuscate their existence in the reported statistic. This is why prediction intervals are presently the preferred solution concept.

While prediction intervals avoid some of the unintuitive pitfalls of confidence intervals, it is important to note that the predictive guarantee (2) has subtleties of its own. It is a mixed observational-causal guarantee: coverage does not hold for all $x$, just for most $x$ according to some probability distribution. For example, if we observe only trials with patients younger than 70, then coverage may not hold for those older than 70. Furthermore, the probabilistic guarantee is marginal. For example, if $\alpha = 0.05$, then it is possible for coverage to be 99% for patients younger than 60 and only 80% for patients between 60 and 70, so long as the average is at least 95%.

The guarantee (2) is most reliable when the distribution over $x$ is explicitly specified by a generative model. If trial designs are actually chosen according to this distribution, and $x$ consists solely of prospectively-set, controllable variables, then it is easy to sample future $x$ for which the coverage guarantee holds. If $x$ includes retrospectively-collected information, or the trials are designed according to unspecified criteria, then the guarantee becomes less meaningful.

## 2.3 Randomized Controlled Trials (RCTs)

An RCT enrolls $m$ participants with potential outcomes $\rho_1, \ldots, \rho_m$. Uniformly at random, it assigns $m_0$ of them to group 0 (the comparison), and the remaining $m_1$ to group 1 (the treatment). Most RCTs do not report individual outcomes. Rather, they report the mean and (corrected) variance of the comparison outcomes are reported as $y^{(0)}$ and $v^{(0)}$. The same statistics are reported for the treatment outcomes as $y^{(1)}$ and $v^{(1)}$. These are combined into $y$, the difference in means, and $v$, a sum of the observed standard errors [Deeks and Higgins, 2010]. These statistics are defined as:

$$y^{(g)} = \frac{1}{m_g} \sum_{i \text{ in group } g} \rho_i(g) \qquad\qquad y = y^{(1)} - y^{(0)}$$

$$v^{(g)} = \frac{1}{m_g - 1} \sum_{i \text{ in group } g} (\rho_i(g) - y^{(g)})^2 \qquad\qquad v = \frac{v^{(0)}}{m_0} + \frac{v^{(1)}}{m_1}$$

Condensing the data into $y$ and $v$ has the following rationale. It can be shown that $y$ is an unbiased estimate of the CATE:

$$\mathbb{E}(y \mid x) = \mathbb{E}(u \mid x)$$

Thus, as the RCT enrolls a very large number of participants, $y$ converges to $u$. This is the primary reason why RCTs are so valuable. $v$ is an estimate of $y$'s variance around $u$, under conditions discussed in the next section.

## 2.4 Random-Effects Model of the Data

Meta-analysis is conducted upon $n$ trials, each with data $X_i \in \mathcal{X}$, $Y_i \in \mathbb{R}$ and $V_i > 0$ for $i = 1, \ldots, n$. As discussed above, each trial's $Y_i$ is centered around $U_i$, but varies around it due to its limited number of participants. Because $Y_i$ is a sample average, by the central limit theorem, it is asymptotically normally distributed around $U_i$. The random-effects model of meta-analysis [DerSimonian and Laird, 1986, Higgins et al., 2009] asserts, as a simplifying assumption, that $Y_i$ is exactly (not just asymptotically) normally distributed around $U_i$ with true variance equal to the observed one. That is, $Y_i \sim N(U_i, V_i)$. This can be written in a way that highlights a key difference between the standard random-effects model and this paper's model:

$$Y_i(X_i, \xi_i) = \text{ATE} + \underbrace{U_i(X_i, \xi_i) - \text{ATE}}_{\text{between-trial heterogeneity}} + \underbrace{N(0, V_i)}_{\text{within-trial variation}} \qquad (3)$$

The first and last terms are the same in both models. The random-effects model asserts that the middle term $U_i - \text{ATE} \sim N(0, \nu)$ where $\nu$ (often denoted by $\tau^2$) is called the heterogeneity variance. By contrast, in this paper, $U_i$ depends on the features $X_i$, and may also involve arbitrary (non-Gaussian) noise through $\xi_i$. Thus, this paper eliminates a potentially-unrealistic assumption.

## 2.5 Untrusted Data as a Probability Distribution

Independently of RCTs, practitioners and researchers often possess deep intuitions about the CATE. These intuitions arise from the lower levels of the evidence hierarchy: observational studies, individually-published cases, hands-on experience, and personal belief [Murad et al., 2016]. It is difficult to rigorously infer causation from such untrusted (or "real-world") data, since they are observational and may have deeply-embedded biases. Nonetheless, it is often found that untrusted data agree with RCTs [Benson and Hartz, 2000]. Retrospectively, Toews et al. [2024] found the ratio of risk-ratios between RCTs and observational studies to be approximately 1.08. The prospective RCT-DUPLICATE trial found their Pearson correlation to be 0.82 [Wang et al., 2023], with much of the discrepancy attributable to readily-identified factors [Heyard et al., 2024].

Since untrusted data originates from different kinds of sources and experiences, it does not share the form of RCTs. A modern approach to capturing disparate, large quantities of knowledge is to pretrain foundation models. Such models are already being developed for healthcare [Moor et al., 2023, Singhal et al., 2023, Tu et al., 2024]. Concretely, this learns an embedding $\phi(x)$ which maps features $x$ into a Euclidean space having inner product $\kappa(x, x') = \phi(x)^T \phi(x')$. On top of this embedding, a linear predictor of the CATE can be trained as $\mu(x) = w^T \phi(x)$. Practically, this representation $(\mu, \kappa)$ encompasses nearly every useful way of predicting the CATE. Mathematically, this representation constructs a Gaussian process, a probability distribution over functions $f : \mathcal{X} \mapsto \mathbb{R}$, with higher weight placed on $f$ which could plausibly approximate the CATE [Kanagawa et al., 2018, Williams and Rasmussen, 2006]. In this probabilistic perspective, $\mu(x) = \mathbf{E}_f f(x)$ and $\kappa(x, x') = \mathbf{E}_f (f(x) - \mu(x))(f(x') - \mu(x'))$. Gaussian processes are often used as prior probability distributions in Bayesian inference [Gelman et al., 1995]. In our approach, unlike Bayesian inference, the prior distribution is not assumed to be correct.

An important restriction is that $\mu$ and $\kappa$ are fixed with respect to $\mathbb{P}$. In practical terms, this means the outcomes of the trials are not reincorporated into $\mu$ and $\kappa$. Otherwise, the trials could trivially, erroneously serve as their own reality check. Thus, although $\mu$ and $\kappa$ are completely untrusted in terms of their veracity and utility, their provenance (especially the data used to generate them) must be clearly understood. Practices such as preregistration and data transparency can facilitate this understanding [Munafò et al., 2017]. Importantly, this assumption is about the processes used to include data, which are under our control. It is not about the complex phenomena which generate the data itself. It is much weaker than the assumptions of ignorability and positivity which are made in causal inference.

## 2.6 Standard Meta-Analysis Algorithms

As previously mentioned, prevalent algorithms for meta-analysis ignore the features $x$; in the parlance of the field, they perform mean-effect prediction rather than meta-regression. Thus, they simply return a single prediction interval $C \subset \mathbb{R}$ rather than a prediction band. Because the model (3) is not analytically solvable, there is no exact, rigorous frequentist prediction interval. Instead, there are many different formulae [Veroniki et al., 2019, Nagashima et al., 2021], each involving approximations which hold only as $n \to \infty$. Most of the prediction intervals have this form:

$$C = \widehat{\text{ATE}} \pm t\sqrt{\hat{\nu} + \widehat{\text{Var}(\widehat{\text{ATE}})}} \tag{4}$$

In this expression, the variance estimates $\hat{\nu}$ and $\widehat{\text{Var}}(\widehat{\text{ATE}})$ are usually algorithm-specific. More generally, $t$ is the $1 - \frac{\alpha}{2}$ quantile of a Student $t$ distribution with $n - 1$ degrees of freedom. $\widehat{\text{ATE}}$ is an estimate of ATE, usually based upon inverse-variance weighting:

$$\widehat{\text{ATE}} = \sum_i w_i Y_i \bigg/ \sum_i w_i \qquad \text{where } w_i = \frac{1}{V_i + \hat{\nu}} \text{ for each } i = 1, \ldots, n \qquad (5)$$

In practice, the most widely-used prediction interval is based on the classical heterogeneity estimator $\hat{\nu}$ of DerSimonian and Laird [1986], and an estimator $\widehat{\text{Var}}(\widehat{\text{ATE}})$ proposed by Higgins et al. [2009]. When $n$ is small, experimental evidence indicates this interval is too small to satisfy (2) with the desired probability $1 - \alpha$.

**Proposition 1** (Classical Prediction Interval). *Assume the model (3) with $U_i \sim N(\text{ATE}, \nu)$. Define the following quantities within (4):*

$$\hat{\nu} = \frac{Q - (n-1)}{S_1 + S_2/S_1} \quad \widehat{\text{Var}}(\widehat{\text{ATE}}) = (\sum_i w_i)^2 \quad \bar{Y} = \frac{\sum_{i=1}^n V_i^{-1} Y_i}{\sum_{i=1}^n V_i^{-1}} \quad Q = \sum_{i=1}^n V_i^{-1}(Y_i - \bar{Y})^2 \quad S_r = \sum_{i=1}^n V_i^{-r}$$

*Then $C$, as defined in (4), approximately satisfies (2) as $n \to \infty$.*

Partlett and Riley [2017] proposed an alternative prediction interval based upon restricted maximum likelihood (REML) and Hartung-Knapp-Sidik-Jonkman (HKSJ) estimators [Nagashima et al., 2021]. REML obtains $\hat{\nu}$ and $\widehat{\text{ATE}}$ as the maximizers of a log-likelihood function $\ell(\hat{\nu}, \widehat{\text{ATE}})$ which is filtered to remove influences from irrelevant variables [Viechtbauer, 2005]. It is not concave, so it cannot be maximized by standard algorithms. However, its stationary points $\partial \ell / d\hat{\nu} = 0$ (for fixed $\widehat{\text{ATE}}$) and $\partial \ell / d\widehat{\text{ATE}} = 0$ (for fixed $\hat{\nu}$) have closed-form expressions, so it is amenable to alternating maximization. The following estimator $\widehat{\text{Var}}(\widehat{\text{ATE}})$ was developed independently by Hartung and Knapp [2001] and Sidik and Jonkman [2003]. Cochrane Statistical Methods and other groups endorse the use of HKSJ [IntHout et al., 2014, Veroniki, 2022, Veroniki et al., 2019].

**Proposition 2** (REML+HKSJ Prediction Interval). *Assume the model (3) with $U_i \sim N(\text{ATE}, \nu)$. Initialize $\hat{\nu} = 0$. Alternate the updates to $\widehat{\text{ATE}}$ and $w$ in (5) with the following update of $\hat{\nu}$, until a fixed point is approximately reached:*

$$\hat{\nu} \leftarrow \frac{\sum_{i=1}^n w_i^2((Y_i - \widehat{\text{ATE}})^2 - V_i)}{\sum_{i=1}^n w_i^2} + \frac{1}{\sum_{i=1}^n w_i} \qquad \widehat{\text{Var}}(\widehat{\text{ATE}}) = \sum_{i=1}^n \frac{(Y_i - \widehat{\text{ATE}})^2 w_i}{(n-1) \sum_j w_j}$$

*Then $C$, as defined in (4), approximately satisfies (2) as $n \to \infty$.*

In addition to these frequentist intervals, Bayesian intervals for $u$ can also be obtained [Smith et al., 1995, Gelman et al., 1995]. These begin with prior distributions over ATE and $\nu$. Improper (i.e. unnormalized) uniform priors are a default uninformative choice [Röver, 2017]. Using the random-effects model as a likelihood, Bayes' theorem obtains the posterior distribution over ATE and $\nu$, which induces a (normal) posterior distribution over $u$. From this posterior distribution, a prediction interval for $u$ can be derived. Such intervals can be highly sensitive to the choice of uninformative prior, which is partially why Bayesian methods are less common in systematic reviews [Hamaguchi et al., 2021]. Nonetheless, there are some circumstances where the flexibility of Bayesian methods is desirable. For example, the Bayesian approach can be extended to predicting trials. The posterior distribution for future $y \sim N(u, v)$ is just $u$'s posterior with $v$ more variance.

**Proposition 3** (Bayesian Trial Prediction). *Let the prior distribution over ATE be improper uniform. Assume the likelihood (3) with $U_i \mid ATE, \nu \sim N(ATE, \nu)$. Then, recalling (5), the posterior predictive distribution conditioned on $\nu$ is $y \mid \nu = \hat{\nu} \sim N\left(\widehat{ATE}, (\sum_i w_i)^{-1} + \hat{\nu} + v\right)$. [Röver, 2017]*

## 2.7 The Ethics of Meta-Analysis

Healthcare is important, uncertain, and sometimes controversial. Evidence-based medicine was introduced to help resolve some of these issues, but it involves controversy of its own. It unavoidably privileges certain kinds of experiences and opinions over others. This paper does not introduce these problems, but it does operate in their midst. Let us examine how these problems could be ameliorated or aggravated by our approach.

Currently, meta-analysis in evidence-based medicine is highly exclusionary. The "lower levels" of the evidence hierarchy are deprecated in favor of RCTs in an effort to preserve rigor and eliminate bias. However, this introduces some bias of its own. For example, RCTs are expensive to conduct. Any methodology that substantially prefers RCTs may be substantially influenced by funding agencies and associated institutions [Lundh et al., 2017]. Furthermore, RCTs are not ethical to conduct in many situations [Morris and Nelson, 2007]. Conformal meta-analysis recognizes that RCTs are especially valuable, but it holistically incorporates data of less rarified origin. Even when our methods do not lead to quantitative improvements, they are arguably more fair, inclusive, and comprehensive. They could ameliorate concerns that evidence-based medicine limits the autonomy of healthcare professionals [Armstrong, 2007].

However, conformal meta-analysis introduces additional computational and statistical complexity into the process of meta-analysis. This complexity could be exploited by bad actors, with negative societal consequences. For example, a malicious meta-analyst could sneak RCT data into their prior to arrive at essentially whichever conclusions they desire. To prevent such harms from occurring, any rigorous conclusions derived from conformal meta-analysis need to be accompanied by safeguards on the handling of data.

## 3 Related Work

**Causal inference from observational data**. Performing randomized, controlled trials is not the only way to estimate causal effects. After making appropriate assumptions, causal inferences can be extracted from observational data [Imbens and Rubin, 2015, Pearl, 2009, Spirtes et al., 2001]. This is an extensive research endeavor encompassing fields such as economics, public policy and online advertising; we mention some of the most relevant work here. The survey by Colnet et al. [2024] discusses various approaches to integrating RCTs with observational data. To estimate the CATE, causal forests [Wager and Athey, 2018] and metalearners [Künzel et al., 2019] combine machine learning techniques with causal reasoning. The most widespread assumption of such methods is ignorability, or unconfoundedness. It requires that, having observed the features $x$, the treatment assigned to a participant is independent of their potential outcomes $\rho(0)$ and $\rho(1)$. That is, there are no unmeasured variables outside of $x$ that could bias treatment towards different participants. Another widespread assumption is positivity, or overlap: for every $x$, both the treatment and the comparison have a chance of being assigned.

Such strong, unproven assumptions are plausible in many circumstances, but they are not appropri-ate for systematic reviews. At some point, assumptions must be tested; systematic reviews, more confirmatory than exploratory in nature, often serve this crucial purpose. Nevertheless, conformal meta-analysis allows these methods to be (indirectly) used in systematic reviews, without any concerns about their unproven assumptions. These methods can ideally be used to extract better $\mu$ and $\kappa$ from the untrusted data. Indeed, taming severe biases in this data likely requires aggressive assumptions of some kind. Thus, conformal meta-analysis doesn't replace these methods; rather, it expands their domain of application to more scientific settings.

**Conformal prediction with label noise**. Previous works have examined how to conformally predict the underlying $u$ while observing only noisy $Y_1, \ldots, Y_n$. It is often empirically observed that conformal prediction can be obliviously robust to label noise, in the sense that $C(x)$, without any involvement of $V$ or $v$, manages to covers $u$ without any loss in confidence. However, provable guarantees remain elusive. Feldman et al. [2023] show that if $C(x)$ always contains the median of $u \mid x$, then $C(x)$ covers $u$ with no loss in confidence. This is a very strong assumption in meta-analysis, as it essentially posits that the relationship between $x$ and $u$ has been globally determined, and the main difficulty of conformal prediction is to account for the uncertainty driven by the unobserved variables $\xi$. Most approaches to (non-obliviously) handling noise involve some modification to split conformal prediction. In classification, the (discrete) labels may be noisy because they are the majority vote from some underlying probability distribution, which reflects uncertainty over the true class. Stutz et al. [2023] adapt split conformal prediction to account for this uncertainty by sampling multiple labels from the underlying distribution. Sesia et al. [2023] and Penso and Goldberger [2024] modify split conformal prediction to estimate the amount of over (or under) coverage of $C(x)$. Unfortunately, splitting the data is not feasible in meta-analysis, where $n$ is small. Label noise should be distinguished from label shift, when the training $Y_1, \ldots, Y_n$ are sampled from a different distribution than the test $y$ [Podkopaev and Ramdas, 2021].

**Meta-regression**. A meta-regression fits the observed effects $Y_i$ as a (typically linear) function of the features $X_i$ [Stanley and Jarrell, 1989]. Meta-regression is usually conducted to diagnose which features are responsible for heterogeneity. It can also generate useful hypotheses for future research, by identifying which features are associated with higher or lower effects. While meta-regression and conformal-meta-analysis are similar in form, there are a number of crucial differences. Most importantly, unlike conformal-meta analysis, meta-regression does not offer predictive guarantees for new $x$; the fit to the data is post-hoc and interpretive [Baker et al., 2009, Thompson and Higgins, 2002]. The (non-predictive) statistical task in meta-regression is to determine which features have a statistically significant relationship with the effect [Huizenga et al., 2011]. To limit spurious findings, meta-regression is typically performed on a small number of prespecified features. By contrast, conformal meta-analysis fits powerful, nonlinear models on a potentially large number of features. In conformal meta-analysis, the regression, as embodied by the prediction band $C$, is presented as the main result, not just an adjunct diagnostic.

**Bayesian priors**. Conformal meta-analysis takes a prior probability distribution, along with trial data, and makes predictions from a posterior distribution — a process that mirrors Bayesian inference [Gelman et al., 1995]. A crucial difference is that Bayesian inference assumes the prior to be correct, whereas conformal meta-analysis does not. When Bayesian methods are employed in meta-analysis, only uninformative priors are used for ATE. Even with this limitation, the choice of uninformative prior can seriously impact the empirical validity of the ensuing meta-analysis [Hamaguchi et al., 2021]. However, informative priors for the heterogeneity variance $\nu$ have been developed, and are more widely accepted [Rhodes et al., 2016, Lilienthal et al., 2024].

**Uniform confidence bands**. Prediction intervals also should not be confused with uniform confidence bands, which offer the following stronger guarantee, and do not involve unobserved $\xi$:

$$\mathbb{P}_C \left( \text{for all } x \in \mathcal{X}, \ u(x) \in C(x) \right) \geq 1 - \alpha$$

Such bands have been developed for Gaussian process regression in the context of online optimization, where new points $x$ are sequentially, adaptively chosen to minimize uncertainty about $u$ [Srinivas et al., 2009, Chowdhury and Gopalan, 2017, Fiedler et al., 2021, Neiswanger and Ramdas, 2021]. Since subsequent $x$ are chosen adaptively using the band, it is essential for the band to hold for arbitrary $x$ rather than just randomly-sampled $x$. Strictly speaking, these bands are correct for arbitrary $\mu$ and $\kappa$. However, their widths depend on the smoothness of $u$, as quantified by its norm in the reproducing kernel Hilbert space induced by $\kappa$. Since $u$ is unknown, this quantity is also unknown. As a practical matter, when $\mu$ and $\kappa$ can range from very good to very poor, the band is either very wide or unknown. Though conformal meta-analysis only offers prediction intervals with marginal coverage guarantees, their width and coverage do not depend on unknown quantities.

**Utilizing unlabeled data**. Trusted labels are generally considered a scarce resource in machine learning, especially compared to unlabeled data (i.e. $x$ sampled from the marginal distribution of $\mathbb{P}$). Unlabeled data are commonly used to pretrain large foundation models [Dahl et al., 2011, Dai and Le, 2015]. Semi-supervised learning studies how to rigorously use unlabeled data to improve predictions [Balcan and Blum, 2010]. Angelopoulos et al. [2023] recently proposed prediction-powered inference as an approach to safely tighten confidence intervals by using unlabeled data along with a prior derived from separate, untrusted data. In this approach, (1) the unlabeled data and prior (which is temporarily treated as correct) are used to estimate the parameter, (2) concentration inequalities are applied to bound the estimation error arising from limited unlabeled data, and (3) the labeled data are used to correct the estimation error due to inaccuracy of the prior. Subsequently, Zrnic and Candès [2024] proposed cross-prediction-powered inference, which has similar goals but does not depend upon an untrusted prior. Instead, it splits the data (as in cross-validation) to train a predictor. Such methods have been used to improve out-of-distribution causal inference [Demirel et al., 2024]. However, these methods are not directly applicable to predictive meta-analysis, in which there are no unlabeled data. Furthermore, these methods are designed to produce confidence intervals rather than prediction intervals.

**Safely using untrusted data**. Various endeavors in statistics and machine learning involve making predictions that are rigorously guaranteed, even though they use untrusted data. To some extent, all these techniques manage to circumvent the "garbage-in, garbage-out" principle. PAC-Bayesian generalization theory formalizes inductive bias as an (untrusted) prior probability distribution [Shawe-Taylor and Williamson, 1997, McAllester, 1998, Seeger, 2002]. Its generalization bounds are tight when the prior and data align, so that a learning algorithm (producing a posterior distribution) can fit the data without diverging far from the prior. While PAC-Bayes is a very useful theoretical tool, conformal prediction bounds are quantitatively tighter, especially when $n$ is small. In statistics, an untrusted prior distribution can be used to define an e-value, a nonnegative statistic whose mean is at most one [Neiswanger and Ramdas, 2021]. Using its reciprocal as an unnormalized density leads to e-posteriors, which can be used as the basis for valid inferences and decisions [Grünwald, 2023]. To derive confidence intervals with conditional coverage guarantees, likelihood-free inference methods can exploit untrusted prior information [Masserano et al., 2023]. In computer science, algorithms can be infused with untrusted predictions, also called side information, advice, or hints [Mitzenmacher and Vassilvitskii, 2022]. When the predictions are good, the algorithms run faster; when the predictions are bad, the algorithms retain acceptable worst-case performance.

A prototypical example is binary search, which can be modified to run in $O(1)$ time given a good prediction of the target's index, and in $O(\log n)$ time no matter how bad the prediction was.

## 4  Fully-Conformal Idiocentric Ridge

```python
1   def precomputations(M, K, U, m, k, kₒ, α):
2     n = len(M)
3     I = eye(n)
4     I_ = eye(n+1)
5     M_ = append(M,m)
6     K_ = block([[K, k[:, newaxis]], [k, kₒ]])
7     λ = amax(diag(K_))
8     t_ = solve(K_/λ + I_, M_)
9     Q_ = solve(K_+λ*I_, K_)
10    Q = Q_[:-1,:-1]
11    q = Q_[-1,:-1]
12    qₒ = Q_[-1,-1]
13
14    A = -q
15    a = 1-qₒ
16    B = U - Q@U - t_[:-1]
17    b = -q@U - t_[-1]
18    # a is already positive; flip signs (wlog) so that a,A_i >= 0
19    B *= sign(A) + (A == 0)
20    A *= sign(A)
21    S = sqrt(λ*diag(Q))
22    s = sqrt(λ*qₒ)
23
24    τ = ceil((1-α)*(n+1)).astype(int32)
25    return Q, q, qₒ, t_, A, a, B, b, S, s, λ, n, I, τ
```

Algorithm 1: Python / NumPy code for common linear-algebraic computations described in Section 4. In this code, and the code throughout the paper, some elisions and deoptimizations are made for readability. In particular, import statements are omitted.

This section studies standard, noise-free conformal prediction: given observations $X$ and $U$, we seek a prediction interval $C(x)$ for $u$. We fully conformalize kernel ridge regression (KRR) under the condition of idiocentricity, as defined below. Our approach actually generalizes beyond KRR to all linear smoothers, which includes methods such as $k$-nearest neighbors, Nadaraya-Watson kernel regression, and smoothing splines [Buja et al., 1989]. We obtain a fast, simple algorithm for (approximate) full conformal prediction which enjoys the following theoretical guarantee. Section 5 will use it to develop meta-analysis algorithms.

**Theorem 1** (Fully-Conformal Idiocentric KRR). *Let $(X_i, U_i) \sim \mathbb{P}$ (for $i = 1, \ldots, n$) as well as $(x, u) \sim \mathbb{P}$ be exchangeable. Let $\mu$ and $\kappa$ be fixed with respect to $\mathbb{P}$. Algorithm 2 returns $[u_-, u_+]$ satisfying $\mathbb{P}(u \in [u_-, u_+]) \geq 1 - \alpha$.*

## 4.1 Kernel Ridge Regression, a Linear Smoother

Let $M$ and $K$ be the mean and kernel function applied to the training features:

$$M = [\mu(X_1), \ldots, \mu(X_n)]^T \in \mathbb{R}^n \qquad K = [\kappa(X_i, X_j)]_{1 \leq i,j \leq n} \in \mathbb{R}^{n \times n}$$

This has normalized residuals $|U_i - M_i|/\sqrt{K_{ii}}$ for $i = 1, \ldots, n$. Given a parameter $\lambda \in \mathbb{R}$, KRR learns the following posterior on the training features:

$$\widehat{M} = (\widehat{K}/\lambda)U + (K/\lambda + I)^{-1}M \qquad \widehat{K} = \lambda(K + \lambda I)^{-1}K$$

In full conformal prediction, KRR is applied to the training set $(X, U)$ augmented by $(x, u)$. We will use bars to denote this augmentation, so $\bar{X} = [X; x]$, $\bar{U} = [U; u]$. Let $m = \mu(x)$, $k = [\kappa(X_1, x), \ldots, \kappa(X_n, x)]^T$, $k_0 = \kappa(x, x)$, and:

$$\bar{I} = \begin{bmatrix} I & 0 \\ 0 & 1 \end{bmatrix} \qquad \bar{K} = \begin{bmatrix} K & k \\ k^T & k_0 \end{bmatrix} \qquad \bar{Q} := (\bar{K} + \lambda\bar{I})^{-1}\bar{K} = \begin{bmatrix} Q & q \\ q^T & q_0 \end{bmatrix}$$

Then, the augmented posterior mean is:

$$\begin{bmatrix} \widehat{M} \\ \hat{m} \end{bmatrix} = \bar{Q}\begin{bmatrix} U \\ u \end{bmatrix} + \overbrace{(\bar{K}/\lambda + \bar{I})^{-1}\begin{bmatrix} M \\ m \end{bmatrix}}^{\bar{t} = [t; t_0]}$$

So the differences between the observations and posterior means are:

$$\begin{bmatrix} U - \widehat{M} \\ u - \hat{m} \end{bmatrix} = (\bar{I} - \bar{Q})\begin{bmatrix} U \\ u \end{bmatrix} - \bar{t} = \begin{bmatrix} (I - Q)U - qu \\ -q^T U + (1 - q_0)u \end{bmatrix} - \bar{t} = \begin{bmatrix} Au + B \\ au + b \end{bmatrix}$$

with the abbreviations:

$$\begin{bmatrix} A \\ a \end{bmatrix} = \begin{bmatrix} -q \\ 1 - q_0 \end{bmatrix} \qquad \begin{bmatrix} B \\ b \end{bmatrix} = \begin{bmatrix} I - Q \\ -q^T \end{bmatrix}U - \bar{t}$$

The augmented posterior kernel matrix is $\lambda\bar{Q}$. Therefore the posterior residuals are:

$$R_i = \frac{|A_i u + B_i|}{S_i} \qquad\qquad r = \frac{|au + b|}{s} \qquad (6)$$

with the final abbreviations $S_i = \sqrt{\lambda Q_{ii}}$ and $s = \sqrt{\lambda q_0}$. Conveniently, these residuals are absolute values of affine functions in $u$. It is easy to see that residuals of this form are shared by the following broader class of learning algorithms.

**Definition 1** (Symmetric Linear Smoothers). *A learning algorithm is symmetric if permuting its inputs $(X_i, U_i)$ permutes its training predictions $\widehat{M}_i$. A linear smoother's predictions $\widehat{M}$ are linear functions of the observations $U$, though they may be nonlinear in $X$.*

## 4.2 Full Conformal Prediction Under Idiocentricity

Full conformal prediction is based on the following observation.

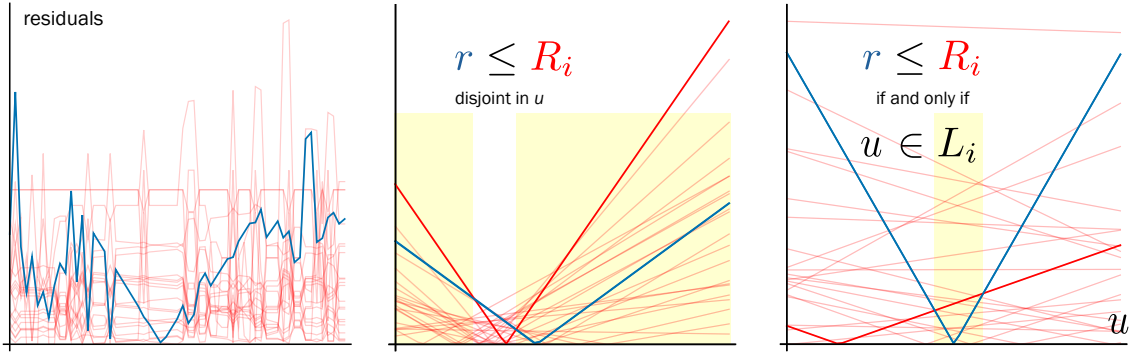

Figure 3: A visualization of how general conformal prediction (left) simplifies for linear smoothers (middle) and even further simplifies under idiocentricity (right). The full conformal interval (7) intersects multiple regions where $r \leq R_i$. In each of the plots, these residuals (vertical axis) are shown as a function of the test example's $u$ (horizontal axis). In general, these regions have no structure and must be determined through exhaustive retraining at every $u$. For linear smoothers (middle), these regions simplify considerably, but may still be disjoint and/or unbounded. Under idiocentricity (right), these regions are just bounded intervals. Geometrically, idiocentricity ensures a larger slope for $r$ than for any $R_i$.

**Proposition 4** (Full Conformal Prediction). *Let $(X_i, U_i) \sim \mathbb{P}$ (for $i = 1, \ldots, n$) as well as $(x, u^*) \sim \mathbb{P}$ be exchangeable. Let $[R; r]$ be the residuals of a symmetric learning algorithm on $[X; x]$ and $[U; u]$. Given any $\alpha \in (0, 1)$, let:*

$$C(x) = \{u : r \text{ is among the } \tau \text{ smallest of } R_1, \ldots, R_n\} \qquad \text{for } \tau = \lceil (1-\alpha)(n+1) \rceil \qquad (7)$$

*Then $\mathbb{P}(u^* \in C(x)) \geq 1 - \alpha$. [Vovk et al., 2005]*

Burnaev and Nazarov [2016] derived an algorithm for computing KRR's $C(x)$. We dramatically simplify the algorithm under the following condition.

**Definition 2** (Idiocentricity). *The residuals $R_1, \ldots, R_n, r$ are idiocentric if:*

$$\lim_{u \to \pm\infty} \frac{|\partial r/\partial u|}{|\partial R_i/\partial u|} > 1 \quad \text{for all } i = 1, \ldots, n$$

This condition means that changing the test example's effect $u$ changes its own residual more than it changes the residuals of other examples. By exchangeability and symmetry, this applies to all the examples, not just the test one. For linear smoothers, idiocentricity simplifies to the following condition.

**Lemma 1** (Linear Idiocentricity). *The residuals in (6) are idiocentric if $\frac{|a|}{s} > \frac{|A_i|}{S_i}$ for all $i$.*

First, let us show how this condition simplifies $C(x)$.

**Lemma 2.** *For $i = 1, \ldots, n$, let $L_i = \left[\frac{-S_i b - s B_i}{S_i a + s A_i}, \frac{-S_i b + s B_i}{S_i a - s A_i}\right]$, noting these endpoints may not be sorted. If KRR is idiocentric, then its prediction set (7) simplifies to:*

$$C(x) = \{u : u \text{ is outside less than } \tau \text{ of the } L_1, \ldots, L_n\}$$

```python
1   def conformal_krr(M, K, U, m, k, ko, α):
2       _,_,_,_,A,a,B,b,S,s,_,n,_,τ = precomputations(M, K, U, m, k, ko, α)
3   
4       if τ <= n:
5           us = vstack([
6               (-S*b - s*B) / (S*a + s*A),
7               (-S*b + s*B) / (S*a - s*A)
8           ])
9           us = sort(us, axis=0)
10          uₚ = sort(us[1])[τ-1]
11          uₙ = flip(sort(us[0]))[τ-1]
12          return uₙ, uₚ
13      else:
14          return -inf, inf
```

Algorithm 2: Python code for fully-conformalized idiocentric kernel ridge regression. It is mildly approximate due to the simplification introduced by Lemma 3. This algorithm can be used for plain regression outside the context of meta-analysis. Presently, it is used as a subroutine for Algorithm 3.

*Proof.* Since the residuals defined in (6) are absolute values, we can flip the signs of $b$ and $B_i$ to standardize on $a, A_i \geq 0$. Since $a/s > A_i/S_i \geq 0$, the condition $r \leq R_i$ is equivalent to $u \in L_i$. □

We slightly loosen the defining condition of $C(x)$ to obtain an even simpler algorithm.

**Lemma 3.** *In the notation of Lemma 2, let $u_+$ be (the minimal value) above $\tau$ of the $L_i$, and let $u_-$ be (the maximal value) below $\tau$ of the $L_i$. Then $C(x) \subseteq [u_-, u_+]$.*

*Proof.* The upper endpoint $u_+$ is met when, for $\tau$ of the $i \in \{1, \ldots, n\}$, we have $u_+ \leq L_i$ or $u_+ \geq L_i$. Ignore the first possibility, which becomes more unlikely as $u_+$ increases, for a potentially looser but nonetheless valid interval. A similar argument justifies $u_-$. □

KRR is idiocentric when $\lambda$ is set sufficiently large. The following upper bound is sometimes loose, but works well throughout this paper. We note that the optimal setting of $\lambda$ for regression may not coincide with the optimal setting for conformal prediction. For example, $\lambda = 0$ (known as interpolation or ridgeless regression) can be a good learning algorithm [Hastie et al., 2022, Liang and Rakhlin, 2020], but it is useless for full conformal prediction, since its residuals are all zero.

**Theorem 2.** *KRR is idiocentric if $\lambda \geq \max_i \bar{K}_{ii} = \max\{K_{11}, \ldots, K_{nn}, k_0\}$.*

*Proof.* Recalling (1), we seek to prove:

$$\frac{|q_i|}{\sqrt{Q_{ii}}} < \frac{|1 - q_0|}{\sqrt{q_0}} \iff \frac{|q_i|}{\sqrt{Q_{ii} \cdot q_0}} < \frac{|1 - q_0|}{q_0}$$

Since $Q$ is positive definite, by the Cauchy-Schwartz inequality:

$$|q_i| = |\langle f_i, f_0 \rangle| \leq ||f_i|| \cdot ||f_0|| = \sqrt{||f_i||^2 \cdot ||f_0||^2} = \sqrt{Q_{ii} \cdot q_0}$$

Thus, it suffices to show that $1 < \frac{1-q_0}{q_0}$, that is, $0 < q_0 < \frac{1}{2}$. Since $\bar{Q}$ is positive definite, $q_0 > 0$ is obvious. To establish $q_0 < \frac{1}{2}$, let us examine the constraints on the last row of $\bar{Q}$. By the original definition of $\bar{Q}$, taking just the last column of $\bar{K}$:

$$\begin{bmatrix} q \\ q_0 \end{bmatrix} = (\bar{K} + \lambda \bar{I})^{-1} \begin{bmatrix} k \\ w \end{bmatrix}$$

Expanding and multiplying by both sides:

$$\left( \begin{bmatrix} K & k \\ k^T & k_0 \end{bmatrix} + \lambda \bar{I} \right) \begin{bmatrix} q \\ q_0 \end{bmatrix} = \begin{bmatrix} k \\ k_0 \end{bmatrix}$$

Expanding again:

$$\begin{bmatrix} K \\ k^T \end{bmatrix} q + \begin{bmatrix} k \\ k_0 \end{bmatrix} q_0 + \lambda \begin{bmatrix} q \\ q_0 \end{bmatrix} = \begin{bmatrix} k \\ k_0 \end{bmatrix}$$

This finally leads to the constraints:

$$(K + \lambda I)q = (1 - q_0)k$$
$$k^T q + \lambda q_0 = (1 - q_0)k_0$$

Inverting the first equation to solve for $q = (1 - q_0)(K + \lambda I)^{-1} k$ and plugging into the second yields:

$$(1 - q_0)k^T (K + \lambda I)^{-1} k + \lambda q_0 = (1 - q_0)k_0$$

If we take $\lambda = k_0$ then:

$$(1 - q_0)k^T (K + k_0 I)^{-1} k = (1 - 2q_0)k_0$$

$$\sum_{i=1}^{n} \frac{\tilde{k}_i^2}{\lambda_i + k_0} = \frac{1 - 2q_0}{1 - q_0} k_0$$

The left hand side is positive, so in order for the right hand to be positive, it is necessary that $q_0 < \frac{1}{2}$, as originally desired. To ensure $\lambda$ (and KRR overall) remain symmetric, this analysis must be applied to any permutation of the data. Thus, $\lambda$ should be larger than any diagonal entry of $\bar{K}$, not just $k_0$. □

To prove Theorem 1, use the $\lambda$ of Theorem 2 to earn the simplified interval of Lemma 3, which is supported by the coverage guarantee of Proposition 4.

## 5  Conformal Meta-Analysis Algorithms

The main results of this section are the following algorithms for predicting causal effects.

**Theorem 3** (Conformally Predicting Clean Effects). *Let $\delta > 0$. Under the assumptions for Predicting Effects, Algorithm 3 returns $[u_-, u_+]$ satisfying $\mathbb{P}(u \in [u_-, u_+]) \geq (1 - \alpha)(1 - \delta)$.*

**Theorem 4** (Conformally Predicting Effects). *Let $\eta > 0$. Under the assumptions for Predicting Effects, Algorithm 5 returns $[u_-, u_+]$ satisfying $\mathbb{P}(u \in [u_-, u_+]) \geq 1 - \frac{\alpha}{(1-\alpha)\mathrm{erfc}\sqrt{\eta/2}}$.*

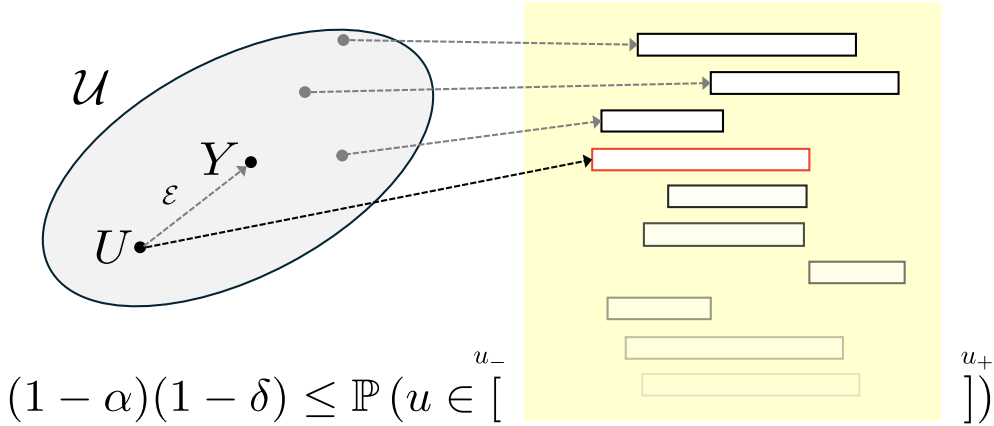

Figure 4: A high-level illustration of Theorem 3. It determines a region $\mathcal{U}$ which contains $U$ with high probability. Each $\widehat{U} \in \mathcal{U}$ induces a different interval $C(x; \widehat{U})$. By covering them all in an outer interval $[u_-, u_+]$, the conformal coverage guarantee of $C(x; U)$ (highlighted in red) is inherited.

In the second algorithm (Algorithm 5), setting $\eta = 0$ (i.e. disabling noise correction) obtains confidence $\frac{1-2\alpha}{1-\alpha}$, which is just a slight loss from $1 - \alpha$ when $\alpha \approx 0$. (For example, 0.95 confidence drops to 0.9473, which probably doesn't change $\tau = \lceil (1-\alpha)(n+1) \rceil$). This setting is appropriate when $V \approx 0$, i.e. the trials all have a large number of participants. By setting $\eta = 2 \cdot \text{inverfc}(\frac{1}{c(1-\alpha)})^2$, the confidence drops to $1 - c \cdot \alpha$. More noise correction is conceptually more appropriate when analyzing both small and large trials. However, the loss of confidence means larger $n$ is needed, which may not be a worthwhile tradeoff. Conformal prediction is usable only when $\tau \leq n$; with $c = 2$, a final confidence of 95% requires $n \geq 40$. This is twice the $n$ needed for $\eta = 0$.

While the overhead at $\eta = 0$ is not practically important, it indicates either the second algorithm, or its analysis, are suboptimal. When meta-analysis is very close to regression (i.e. $V \approx 0$), the original $1 - \alpha$ coverage should be smoothly recovered. The first algorithm (Algorithm 3) has this ideal behavior. This algorithm directly builds upon Algorithm 2 and demonstrates there is no inherent overhead in meta-analysis. Furthermore, its guarantee (Theorem 3) still holds in the much more challenging situation where the noise vector $\mathcal{E}$ (in $Y = U + \mathcal{E}$) is not iid normal, but is adversarially chosen within some set $\mathbf{E}$. However, because of this adversarial pessimism, its intervals become loose as $V$ grows above zero. Thus, we recommend Algorithm 5 for practical application.

In addition to these algorithms for predicting effects, Section 5.2 presents an algorithm for predicting trials, i.e. covering $y$ given $x$ and $v$. This is just an adaptation of fully-conformal KRR (Section 4) which takes into account both $V$ and $v$, incorporating them into the residuals to control for noise. This algorithm is developed primarily as a prelude to Algorithm 5, but it is also used in the case study of Section 7.

**Theorem 5** (Conformally Predicting Trials). *Let $\eta > 0$. Under the assumptions for Predicting Trials, Algorithm 4 returns $[y_-, y_+]$ satisfying $\mathbb{P}(y \in [y_-, y_+]) \geq 1 - \alpha$.*

## 5.1 Predicting Clean Effects

The proof of Theorem 3 decouples into a relatively blunt statistical component (Lemma 4) and a more complicated computational one (Lemma 5). It builds upon Theorem 1's guarantee that fully-

```python
1   def predict_clean_effect(M, K, Y, V, m, k, ko, α, δ):
2       # get "plain" conformal interval for y, having observed Y
3       yn, yp = conformal_krr(M, K, Y, m, k, ko, α)
4       # ω is additional endpoint width due to noise
5       # ω -> 0 if V -> 0 quickly enough
6       Q,q,_,_,A,a,_,_,S,s,_,n,I,τ = precomputations(M, K, Y, m, k, ko, α)
7       ρ = chi2.ppf(1-δ, n)
8       G = vstack([
9           (-S*q[newaxis, :] + s*(I-Q)) / (S*a + s*A),
10          (-S*q[newaxis, :] - s*(I-Q)) / (S*a - s*A) ])
11      ω = amax(norm(G * sqrt(ρ*V), axis=1))
12      # return widened interval
13      un, up = yn-ω, yp+ω
14      return un, up
```

Algorithm 3: Python code for conformal meta-analysis with nearly no noise (i.e. "clean" effects observed in very large trials). As the noise vanishes (i.e. $V \to 0$), this algorithm reduces to Algorithm 2 because $\omega \to 0$.

conformal KRR's interval covers $u$. Let $C(x; \widehat{U})$ denote this interval when $\widehat{U}$ is given as training data. If some outer interval $\widehat{C}(x)$ contains all $C(x; \widehat{U})$ over a plausible set of $\widehat{U}$ including $U$, then of course $\widehat{C}(x)$ contains $C(x; U)$ and inherits its coverage. Lemma 4 shows the uncertainty over $U$ falling in that plausible set separates from fully-conformal KRR's uncertainty over $u$, given $U$. This follows from $\mathcal{E}$'s independence from $U$ given $V$.

**Lemma 4** (Cover All Possibilities). *Let $C(x; U) = C(x)$ be the KRR interval from Lemma 2 when computed on the true $U$. Let $\widehat{C}(x)$ contain all intervals induced by the ellipsoid $\mathbf{E}$:*

$$\mathbf{E} = \left\{ E : \sum_{i=1}^{n} \frac{E_i^2}{V_i} \leq \rho \right\} \qquad \widehat{C}(x) = \bigcup_{E \in \mathbf{E}} C(x; \underbrace{U + \mathcal{E} - E}_{Y})$$

*Let $\rho > 0$ be chosen so that $\mathbb{P}_{\mathcal{E}}(\mathcal{E} \in \mathbf{E} \mid V) \geq 1 - \delta$. Then $\mathbb{P}(u \in \widehat{C}(x)) \geq (1-\alpha)(1-\delta)$.*

*Proof.* In the following, let rest denote $X, U, x, u$.

$$\begin{aligned}
\mathbb{P}(u \in \widehat{C}(x)) &\geq \mathbb{P}(u \in C(x), C(x) \subseteq \widehat{C}(x)) && \text{(partial probability)} \\
&= \mathbb{E}_V \mathbb{E}_{\text{rest}} \left( \mathbf{1}(u \in C(x)) \cdot \mathbb{P}_{\mathcal{E}}(C(x) \subseteq \widehat{C}(x) \mid V, \text{rest}) \right) && \text{(total probability)} \\
&\geq \mathbb{E}_V \mathbb{E}_{\text{rest}} \left( \mathbf{1}(u \in C(x)) \cdot (1-\delta) \right) && \text{(see below)} \\
&= (1-\delta) \mathbb{P}_{\text{rest}, V}(u \in C(x)) && \text{(total probability)} \\
&\geq (1-\delta)(1-\alpha) && \text{(conformal prediction)}
\end{aligned}$$

A sufficient condition for $C(x; U) \subseteq \widehat{C}(x)$ is that $\mathcal{E} = E$ for some $E \in \mathbf{E}$, i.e. that $\mathcal{E}$ belongs to the ellipsoid. Note that $\widehat{C}(x)$ depends on $U$ but this condition does not. Thus:

$$\begin{aligned}
\mathbb{P}_{\mathcal{E}}(C(x) \subseteq \widehat{C}(x) \mid V, \text{rest}) &\geq \mathbb{P}_{\mathcal{E}}(\mathcal{E} \in \mathbf{E} \mid V, \text{rest}) && \text{(sufficient condition)} \\
&= \mathbb{P}_{\mathcal{E}}(\mathcal{E} \in \mathbf{E} \mid V) && \text{(conditional independence)}
\end{aligned}$$

$$\geq 1 - \delta \qquad \text{(assumption)}$$

This lemma doesn't make any smoothness assumptions on how $C(x; \widehat{U})$ changes as $\widehat{U}$ varies away from $U$; it relies on the coverage of exactly $C(x; U)$, but not of any slight perturbation $C(x; \widehat{U})$. Furthermore, the lemma does not strongly depend on the distribution of $\mathcal{E}$, just that we know a set **E** which captures it with probability $1 - \delta$. For Gaussian noise, this is an ellipsoid. □

The previous lemma converts the statistical problem of covering $u$ into the purely computational problem of bounding the endpoints of $\widehat{C}(x)$. The next lemma shows that we can just extend the "naive" interval, which plugs in $Y$ for $U$, with some additional width $\omega$ at the endpoints. $\omega$ is easy to compute, and drops to zero if $V$ (and $\delta$) go to zero. This indicates the algorithm retains optimal confidence as meta-analysis becomes close to plain regression.

**Lemma 5** (Extra Width). *Let $[y_-, y_+]$ be the interval returned by Algorithm 2 when given $Y$ in place of $U$. Let $\rho > 0$ and $G \in \mathbb{R}^{2n \times n}$ be defined by Lemma 4 and (10), respectively. Let $\omega$ be the maximum $\ell_2$ norm of any row of $G \operatorname{diag}\sqrt{\rho V}$. Then $\widehat{C}(x) \subseteq [y_- - \omega, y_+ + \omega]$.*

As the first step of proving this lemma, we allow two separate optimizations for the upper and lower endpoints of $\widehat{C}(x)$. The following lemma performs this (trivially valid) split.

**Lemma 6.** *Recall the intervals $L_i$ from Lemma 2. $\widehat{C}(x) \subseteq [\hat{u}_-, \hat{u}_+]$ where:*

$$\hat{u}_- := \min_{E \in \mathbf{E}} \max\{\text{bottom } n - \tau + 1 \text{ lower endpoints of } L_1, \ldots, L_n\} \tag{8}$$

$$\hat{u}_+ := \max_{E \in \mathbf{E}} \min\{\text{top } n - \tau + 1 \text{ upper endpoints of } L_1, \ldots, L_n\} \tag{9}$$

The $2n$ endpoints (both lower and upper) of the $L_i$ can be expressed as $F + GE$, which highlights their linear dependence on the noise $E$.

**Lemma 7.** *The $2n$ endpoints of $L_1, \ldots, L_n$ are the entries of $F + GE$, where $F \in \mathbb{R}^{2n}$ and $G \in \mathbb{R}^{2n \times n}$ are defined below in (10). (Thus, if $E = 0$, then the bottom and top $n - \tau + 1$ endpoints in $F$ are exactly $\{y_-, y_+\}$ returned by Algorithm 2 when given $Y$ in place of $U$).*

*Proof.* The $L_i$ are defined in terms of equations from Section 4 and Lemma 2, which we summarize here. In terms of the variables $B, b, \widehat{U}$ and $E$ and constants $a, A, s, S, Q, q, \bar{t}, Y$ and $V$:

$$L_i = \left[\frac{-S_i b - s B_i}{S_i a + s A_i}, \frac{-S_i b + s B_i}{S_i a - s A_i}\right] \quad \text{where} \quad \begin{bmatrix} B \\ b \end{bmatrix} = \begin{bmatrix} I - Q \\ -q^T \end{bmatrix} \widehat{U} - \bar{t} \quad \text{and} \quad \widehat{U} = Y - E$$

To simplify notation, we can eliminate the variables $B, b$, and $\widehat{U}$ the endpoints directly in terms of $E$. Specifically, $L = [F^{(0)} + G^{(0)} E, F^{(1)} + G^{(1)} E]$, where:

$$\begin{aligned}
F_i^{(0)} &= \frac{S_i q^T Y + S_i t_0 - s(I - Q) Y + st}{S_i a + s A_i} & G_i^{(0)} &= \frac{-S_i q^T + s(I - Q)}{S_i a + s A_i} \\
F_i^{(1)} &= \frac{S_i q^T Y + S_i t_0 + s(I - Q) Y - st}{S_i a - s A_i} & G_i^{(1)} &= \frac{-S_i q^T - s(I - Q)}{S_i a - s A_i} \\
F &= [F^{(0)}; F^{(1)}] & G &= [G^{(0)}; G^{(1)}]
\end{aligned} \tag{10}$$

□

Let us focus on upper bounding $\hat{u}_+$ from (9); the same proof technique will apply to lower bounding $\hat{u}_-$ from (8). The maximization of $\hat{u}_+$ involves two aspects: choosing $n - \tau + 1$ coordinates (corresponding to large values in $F$) and then optimizing $E$ so $F + GE$ is large in those coordinates. The program is difficult because the largest coordinates of $F$ may not coincide with the largest obtainable by $GE$ — for example, it might be worthwhile to choose an initially large $F_i$ even if $(GE)_i$ cannot become very large. We will allow the adversary to decouple these choices, allowing them to add the largest-possible $(GE)_i$ to the largest coordinates of $F$. This loosens the interval, but greatly simplifies it as well.

**Lemma 8.** *Let $y_+$ be the upper endpoint returned by Algorithm 2 when given $Y$ in place of $U$. Recall the definition of $\hat{u}_+$ in (9). Then $\hat{u}_+ \leq u_+ := y_+ + \omega$, where:*

$$\omega := \max_E \max_i (GE)_i \text{ such that } E \in \mathbf{E}$$

*Proof.* As discussed above, $\hat{u}_+$ is produced by taking some initial upper endpoint $F_i$ and choosing $E$ to maximize $(GE)_i$. $y_+$ is the top-$(n - \tau + 1)$ endpoint, so $y_+ + \omega = u_+$ exceeds most of the possible $F + (GE)_i$: it is not possible for any initially-smaller endpoint $F_i \leq y_+$ to become larger than $u_+$ following the introduction of noise, since then $F_i + (GE)_i \leq F_i + \omega \leq y_+ + \omega = u_+$. Even at the $n - \tau$ initially-larger endpoints $F_i \geq y_+$, it may be possible for $u_+$ to exceed $F_i + (GE)_i$. But this just means the upper bound is looser than necessary. □

Now, the excess width $\omega$ can be easily computed with basic linear algebra. The following lemma concludes the proof of Lemma 5.

**Lemma 9.** *$\omega$ is the maximum $\ell_2$ norm of any row of $G \operatorname{diag}\sqrt{\rho V}$.*

*Proof.* Because $\mathbf{E}$ is symmetric around the origin, an absolute value can be introduced to the definition of $\omega$ without affecting its value, which makes it an $\ell_\infty$ norm:

$$\omega = \max_E ||GE||_\infty \text{ such that } E \in \mathbf{E}$$

Recalling that $\mathbf{E}$ is an ellipsoid, perform a change of variables to transform it to a unit $\ell_2$ ball, yielding the following equivalent program:

$$\omega = \max_E ||G \operatorname{diag}(\sqrt{\rho V}) E||_\infty \text{ such that } ||E||_2 \leq 1$$

This is the definition of the $2 \to \infty$ operator norm, which is computed as described in the lemma statement. A different geometry for $\mathbf{E}$ would result in a different norm bound on $h$. For example, box constraints for $\mathbf{E}$ would result in $h$ being bounded by the $\infty \to \infty$ norm of $G \operatorname{diag}\sqrt{\rho V}$, i.e. the maximum $\ell_1$ norm of any row. Thus, this simple proof technique (much like Lemma 4) can be extended to different kinds of noise. □

Finally, we need to calculate the threshold $\rho$, in terms of $\delta$ and $V_i$, so that $\mathcal{E} \in \mathbf{E}$ with high probability. By a change of variables to standard normals, $\rho$ equals the $1 - \delta$ quantile of the chi-square distribution with $n$ degrees of freedom. This completes the proof of Theorem 3.

```python
1   # capital arguments for training trials, lowercase are for test
2   def predict_trial(M, K, Y, V, m, k, ko, v, α, η):
3     # standard linear algebra for KRR
4     Q,q,_,_,A,a,B,b,S,s,_,_,I,τ = precomputations(M, K, Y, m, k, ko, α)
5
6     if τ <= n: # enough training trials for conformal prediction
7       # compute interval L_i = G_i±H_i for each training trial
8       # y in L_i corresponds to r <= R_i for residuals
9       D = square(I-Q) @ V
10      d = square(q) @ V
11      S2, s2 = square(S), square(s)
12      a2A2 = a**2*S2 - A**2*s2
13      ρ = η*(D*s2 - d*S2 - a2A2*v)
14      G = (A*B*s2 -a*b*S2) / a2A2
15      H = sqrt(maximum(0, s2*S2*(A*b - a*B)**2 - ρ*a2A2)) / a2A2
16      Lₙ, Lₚ = G-H, G+H
17      # return extreme quantiles of L_i's upper/lower endpoints
18      yₚ = sort(Lₚ)[τ-1]
19      yₙ = flip(sort(Lₙ))[τ-1]
20      return yₙ, yₚ
21    else: # not enough training trials
22      return -inf, inf
```

Algorithm 4: Python code for conformal prediction of trials. This is a variant of Algorithm 2 which incorporates terms correcting for noise in the observed effects.

## 5.2 Predicting Trials

This section proceeds analogously to Section 4, deriving residuals for full conformal prediction. However, now the data are of the form $(x, y, v)$ rather than $(x, u)$. Here is a version of Proposition 4 adapted to the meta-analytic data.

**Proposition 5** (Full Conformal Prediction). *Let $(X_i, Y_i, V_i) \sim \mathbb{P}$ (for $i = 1, \ldots, n$) as well as $(x, y^*, v) \sim \mathbb{P}$ be exchangeable. Let $[R; r]$ be the residuals of a symmetric learning algorithm on $[X; x]$, $[Y; y]$ and $[V; v]$. Given any $\alpha \in (0, 1)$, let $\tau = \lceil (1-\alpha)(n+1) \rceil$ and:*

$$C(x, v) = \{y : r \text{ is among the } \tau \text{ smallest of } R_1, \ldots, R_n\} \quad (11)$$

*Then $\mathbb{P}(y^* \in C(x, v)) \geq 1 - \alpha$. [Vovk et al., 2005]*

Now that $V$ and $v$ are observed, noise-correcting terms based upon them can be incorporated into the residuals. In the following residuals, $A_i, a, B_i, b, S_i,$ and $s$ are the same as in Section 4. For $\eta > 0$, $Z_i$ and $z$ subtract off an $\eta$ multiple of the expected impact of the noise. To simplify the calculation of these expectations, we adopt the squared loss in lieu of the absolute loss:

$$R_i = \frac{(\widehat{M}_i - Y_i)^2 - \eta Z_i}{S_i^2} = \frac{(A_i y + B_i)^2 - \eta Z_i}{S_i^2} \qquad r = \frac{(\hat{m} - y)^2 - \eta z}{s^2} = \frac{(ay + b)^2 - \eta z}{s^2}$$

To determine $Z_i$ and $z$, decompose the differences between the observations and the posterior means. As before, denote augmentation with overlines, as in $\bar{\mathcal{E}} = [\mathcal{E}; \epsilon]$.

$$\begin{bmatrix} Y - \widehat{M} \\ y - \hat{m} \end{bmatrix} = (\bar{I} - \bar{Q})(\bar{U} + \bar{\mathcal{E}} - \bar{M}) - \bar{t}$$

$$= \begin{bmatrix} U - \widehat{M} \\ u - \hat{m} \end{bmatrix} - (\bar{I} + \bar{Q})\bar{\mathcal{E}} = \begin{bmatrix} U - \widehat{M} \\ u - \hat{m} \end{bmatrix} + \begin{bmatrix} (I - Q)\mathcal{E} - q\epsilon \\ -q^T \mathcal{E} + (1 - q_0)\epsilon \end{bmatrix}$$

Now, calculate the mean squared error with respect to $\mathcal{E}_i \sim N(0, V_i)$ and $\epsilon \sim N(0, v)$:

$$\mathbb{E}\,(Y_i - \widehat{M}_i)^2 = \mathbb{E}\,(U_i - \widehat{M}_i + (e_i - Q_i)^T \mathcal{E} - q_i \epsilon)^2$$

$$= (U_i - \widehat{M}_i)^2 + \mathbb{E}\left((1 - Q_{ii})\mathcal{E}_i - \sum_{j \neq i} Q_{i,j} \mathcal{E}_j - q_i \epsilon\right)^2$$

$$= (U_i - \widehat{M}_i)^2 + \underbrace{\overbrace{(1 - Q_{ii})^2 V_i + \sum_{j \neq i} Q_{i,j}^2 V_j}^{Z_i} + \underbrace{q_i^2}_{A_i^2} v}_{D_i}$$

$$\mathbb{E}\,(y - \hat{m})^2 = \mathbb{E}\,(u - \hat{m} - q^T \mathcal{E} + (1 - q_0)\epsilon)^2 = (u - \hat{m})^2 + \underbrace{\sum_j q_j^2 V_j}_{d} + \overbrace{\underbrace{(1 - q_0)^2}_{a^2} v}^{z}$$

As before, we use idiocentricity to simplify the residual comparisons.

**Theorem 6.** *For $i = 1, \ldots, n$, let $\rho_i = \eta(Z_i s^2 - z S_i^2)$. Define the intervals $L_i = G_i \pm H_i$, where:*

$$G_i = \frac{A_i B_i s^2 - ab S_i^2}{(aS_i)^2 - (A_i s)^2} \quad \text{and} \quad H_i = \frac{\sqrt{\max\left(0,\ s^2 S_i^2 (A_i b - a B_i)^2 - \rho_i((aS_i)^2 - (A_i s)^2)\right)}}{(aS_i)^2 - (A_i s)^2}$$

*If KRR is idiocentric, then its full conformal prediction set (11) simplifies to:*

$$C(x, v) = \{y : y \text{ is outside less than } \tau \text{ of the } L_1, \ldots, L_n\}$$

*Proof.* $r \leq R_i$ rewrites to $S_i^2(ay + b)^2 + \rho_i \leq s^2(A_i y + B_i)^2$. Under idiocentricity, which ensures $a/s > A_i/S_i \geq 0$, this is equivalent to $y \in L_i$. □

To prove Theorem 5, use the $\lambda$ of Theorem 2 to earn the simplified interval of Theorem 6, which is supported by the coverage guarantee of Proposition 5.

### 5.3 Predicting Effects

Theorem 5 guarantees that $C(x, v)$ usually covers $y \sim N(u, v)$. We will use this guarantee to derive intervals $C(x)$ that usually cover $u$. We don't have a $v$ to plug into $C(x, v)$, so we have to dig into how $C(x, v)$ works. The claim of Theorem 4 is that $C(x, 0)$ covers $u$ just slightly less often than it covers $y$, so long as the level of noise correction $\eta$ is not too high. This holds because of two counterbalancing properties of $C(x, v)$ that hold for all $v \geq 0$.

```
1  def predict_effect(M, K, Y, V, m, k, ko, α, η):
2      return predict_trial(M, K, Y, V, m, k, ko, 0, α, η)
```

Algorithm 5: Python code for conformal meta-analysis with small trials, deferring entirely to Algorithm 4.

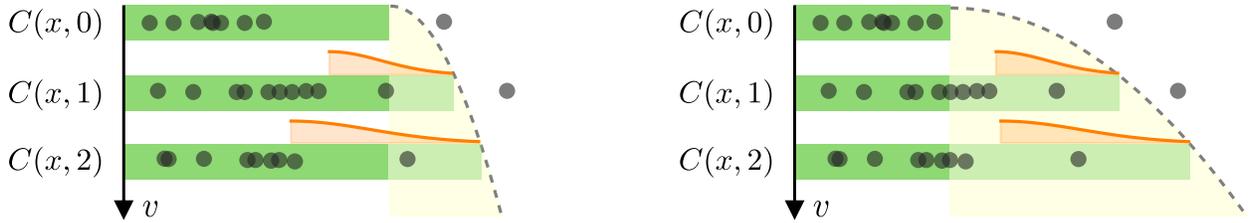

Figure 5: A high-level sketch of $C(x,0)$'s coverage of $u$, when $\eta$ is small enough (left) versus too large (right). The gray dots are $u$, and its distributions conditioned on various $v$ are shown. $C(x,0)$ is the dark green bar; as $v$ increases, $C(x,v)$ increases by $\sqrt{\eta v}$, and that growth (in yellow) is shaved. The orange curves convey the spread of $|N(0,v)|$. With good $\eta$ (left), $C(x,v)$ grows slowly compared to $|N(0,v)|$, which naturally pushes in the $u$ (on average) as $v$ increases. Thus, $C(x,0)$ is wide enough to contain most of the $u$, no matter what $v$ is. On the right, when $\eta$ is large, $C(x,v)$ adapts more dynamically to $v$, so $C(x,0)$ is smaller. Too many $u$ in the yellow region are shaved.

The first property is that most of the spread of $|N(0,v)|$ can be shaved from the edges of $C(x,v)$ without losing too many $u$. This is possible because, in meta-analysis, we care only about small $\alpha$, ideally around 0.05. Since $C(x,v)$ covers $y$ with high probability, there are only a few $u$ closer than $|N(0,v)|$ to the ends of $C(x,v)$ — otherwise, bad flips of the noise could push too many $y$ out of the interval, which would violate the coverage guarantee of $C(x,v)$. While this logic indicates shaving is a conceptually feasible strategy, it remains an abstract possibility, since we don't know $v$, and don't know how much to shave. (It should intuitively be $O(\sqrt{v})$, but constants matter).

The second property is that making $\eta$ smaller limits the growth of $C(x,v)$. We mean this in a completely formulaic sense — we have reasonably concrete expressions for the endpoints of $C(x,v)$, and the following Lemma 10 shows they widen by $\sqrt{\eta v}$. When $\eta = 0$, $C(x,v)$ doesn't depend on $v$ at all. In other words, when noise correction is disabled, $C(x,0)$ must completely internalize the impact of noise, yielding a relatively wide interval. Larger settings of $\eta$ allow $C(x,v)$ to grow more with $v$, allowing (relatively) thin intervals at small $v$. To concretely realize the shaving strategy, we just need to set $\eta$ small enough so that, as a function of $v$, *the shaveable region within $C(x,v)$ grows as fast as $C(x,v)$ itself*. This allows us to obliviously use the baseline $C(x,0)$. The conditional distribution $v \mid x$ is arbitrary and unknown, but any probability mass on $v > 0$ simply pushes more $u$ within $C(x,0)$.

The fact that $C(x,v)$ grows proportionally to $\sqrt{v}$ to capture the noise is not only intuitive, it is necessary. Most well-behaved learning algorithms should yield conformal intervals which grow (on average) at roughly this rate. Our ability to prove an exact growth rate, in the next lemma, relies on the simplicity of full conformal prediction for idiocentric linear smoothers.

**Lemma 10** (Normal Interval Growth). *Let $C(x,v)$ be the interval from Theorem 6. For all $\eta \geq 0$ and $v > 0$, $C(x,v) \subseteq C(x,0) \pm \sqrt{\eta v}$.*

*Proof.* The interval for $y$ depends on $v$ only through $\rho_i$:

$$\frac{1}{\eta}\rho_i = Z_i s^2 - z S_i^2 = D_i s^2 - d S_i^2 - \overbrace{\left((a S_i)^2 - (A_i s)^2\right)} v$$

Under idiocentricity, $a/s > A_i/S_i$. Thus, the bracketed term above is positive, $\rho_i$ decreases with $v$, the square-root radius in $L_i$ (which subtracts $\rho_i$) increases with $v$, and the denominator in $L_i$ is positive. Dividing by the denominator, the radius $H_i$ is of the form $\sqrt{\ldots + \eta v} \leq \sqrt{\ldots} + \sqrt{\eta v}$. Neither the center $G_i$ of $L_i$ nor the other elided terms in the radius depend on $v$; the $\sqrt{\eta v}$ term is the only one which involves $v$. $\square$

The rest of the proof of Theorem 4 doesn't depend on either idiocentricity or linear smoothers. Lemma 11 formalizes the first property described above: most $u$ are contained within $C(x,v)$ by a margin that grows with $v$. Finally, Lemma 12 shows that $C(x,v)$ can be shaved down to $C(x,0)$, with $\eta$ determining the loss in coverage of $u$.

**Lemma 11** (Pay For Room). *Recall $y = u + \epsilon$ for $\epsilon \sim N(0,v)$. Let $w = [u - \epsilon, u + \epsilon]$, with possibly unsorted endpoints. If $\mathbb{P}(y \in C(x,v)) \geq 1 - \alpha$, then $\mathbb{P}(w \subseteq C(x,v)) \geq (1 - 2\alpha)/(1 - \alpha)$.*

*Proof.* Abbreviate $C(x,v) = C$. The key property we repeatedly use is that $y$ is one of the endpoints of $w$ chosen uniformly at random, conditionally independent of the other data. If $w \not\subseteq C$, then either both of its endpoints are not in $C$, or exactly one of them isn't. In the former case, $y$ clearly isn't in $C$; in the latter, it isn't with probability $\frac{1}{2}$. Let gray be the event that exactly one of $w$'s endpoints is outside of $C$. First, we prove that:

$$\mathbb{P}(\mathsf{gray}) \leq 2\alpha \tag{12}$$

Let near denote both of $w$'s endpoints are in $C$, and far that neither are in $C$, so that near, gray, far partition the probability space. By total probability, and the aforementioned reasoning about $y$:

$$\mathbb{P}(y \in C) = (1 - \mathbb{P}(\mathsf{gray}) - \mathbb{P}(\mathsf{far}))\mathbb{P}(y \in C \mid \mathsf{near}) + \mathbb{P}(\mathsf{far})\mathbb{P}(y \in C \mid \mathsf{far}) + \mathbb{P}(\mathsf{gray})\mathbb{P}(y \in C \mid \mathsf{gray})$$

$$= (1 - \mathbb{P}(\mathsf{gray}) - \mathbb{P}(\mathsf{far}))(1) + \mathbb{P}(\mathsf{far})(0) + \mathbb{P}(\mathsf{gray})\frac{1}{2}$$

$$\leq 1 - \mathbb{P}(\mathsf{gray}) + \mathbb{P}(\mathsf{gray})\frac{1}{2}$$

Combining this with the assumption yields (12). Next:

$$\mathbb{P}(y \in C \mid w \not\subseteq C) = \mathbb{P}(\mathsf{gray})\mathbb{P}(y \in C \mid \mathsf{gray}) \qquad \text{(only nonzero case)}$$

$$= \mathbb{P}(\mathsf{gray})\frac{1}{2} \qquad \text{(symmetry)}$$

$$\leq \alpha \qquad \text{(12)}$$

With this inequality, the original claim follows from:

$$1 - \alpha \leq \mathbb{P}(y \in C) \qquad \text{(assumption)}$$

$$= \mathbb{P}(w \subseteq C, y \in C) + (1 - \mathbb{P}(w \subseteq C))\mathbb{P}(y \in C \mid w \not\subseteq C) \qquad \text{(total probability)}$$

$$\leq \mathbb{P}(w \subseteq C, y \in C) + (1 - \mathbb{P}(w \subseteq C))\alpha \qquad \text{(proved above)}$$

$$= \mathbb{P}(w \subseteq C) + (1 - \mathbb{P}(w \subseteq C))\alpha \qquad (y \in w)$$

Note this proof required $\epsilon$ to be symmetric, zero mean, and conditionally independent given its variance $v$, but not necessarily normally distributed. $\square$

**Lemma 12** (Shaving). *If $\mathbb{P}(w \subseteq C(x,v)) \geq \frac{1-2\alpha}{1-\alpha}$, then $\mathbb{P}(u \in C(x,0)) \geq 1 - \frac{\alpha}{(1-\alpha)\mathrm{erfc}\sqrt{\eta/2}}$.*

*Proof.* Abbreviate $C = C(x,v)$ and $\widetilde{C} = C(x,0)$. For the first inequality of the following block, the worst case is obtained when $u$ is exactly one of the endpoints of $\widetilde{C}$ (say, the upper endpoint $\tilde{c}_+$), since that maximizes the distance from the endpoint of $C$ (say, $c_+$), and therefore maximizes probability that $w$ will still remain within $C$.

$$\begin{aligned}
\mathbb{P}(w \subseteq C \mid u \notin \tilde{C}) &\leq \mathbb{P}(\tilde{c}_+ + |\epsilon| \leq c_+) \\
&= \mathbb{P}(|\epsilon| \leq \sqrt{\eta v}) && \text{(Lemma 10)} \\
&= \mathrm{erf}\sqrt{\frac{\eta}{2}} && \text{(normal distribution)}
\end{aligned}$$

Thus, the desired claim follows from total probability and some rearranging:

$$\begin{aligned}
\frac{1-2\alpha}{1-\alpha} &\leq \mathbb{P}(w \subseteq C) && \text{(assumption)} \\
&= \mathbb{P}(u \in \widetilde{C})\mathbb{P}(w \subseteq C \mid u \in \widetilde{C}) + \mathbb{P}(u \notin \widetilde{C})\mathbb{P}(w \subseteq C \mid u \notin \widetilde{C}) && \text{(total probability)} \\
&\leq \mathbb{P}(u \in \widetilde{C}) + (1 - \mathbb{P}(u \in \widetilde{C}))\mathrm{erf}\sqrt{\eta/2} && \text{(proved above)} \quad \square
\end{aligned}$$

## 6 Simulations

We performed four types of simulations on three biomedical datasets from the Penn Machine Learning Benchmark [Olson et al., 2017]. These regression datasets define $K$ and $Y$; we generated synthetic $M$ and $V$ according to parameters prior error $\geq 0$ and effect noise $\geq 0$, respectively. These paramters are formally defined in Appendix A.1. prior error of 0.2, 0.9, and 3.0 roughly correspond to good, okay, and bad priors derived from untrusted data. effect noise between roughly 0 and 10 captures the practical range of large to small trials, but we may use larger values to stress-test algorithms. CMA abbreviates Algorithm 5 with $\eta = 0$. We compare it to the state-of-the-art HKSJ method, which is described in Section 2.6.

**Simulation 1**: This investigates when conformal meta-analysis is superior to traditional meta-analysis. For different settings of prior error, we compare the widths of the intervals obtained by different meta-analysis algorithms. The only situation in which HKSJ is competitive with conformal meta-analysis is when the prior is bad and the number of trials is small/moderate. Otherwise, conformal meta-analysis is superior, sometimes achieving intervals that are dramatically thinner than those of HKSJ.

**Simulation 2**: This experiment checks whether the desired 95% confidence level is still achieved as effect noise increases. Conformal meta-analysis succeeds, whereas HKSJ fails badly. On the other datasets, HKSJ sometimes drops below 80% confidence. This deficiency is present at all settings of effect noise, though it aggravates at higher values. This simulation shows that conformal meta-analysis has a rigorous coverage guarantee, and HKSJ does not. It should be noted that HKSJ was developed to improve the coverage guarantee of the more prevalent Higgins-Thompson-Spiegelhalter method.

**Simulation 3**: This experiment compares different instantiations of Algorithm 5: one with $\eta = 0$, and the other with $\eta = 0.4015$, with $\alpha$ adjusted so both ultimately seek a 90% confidence level.

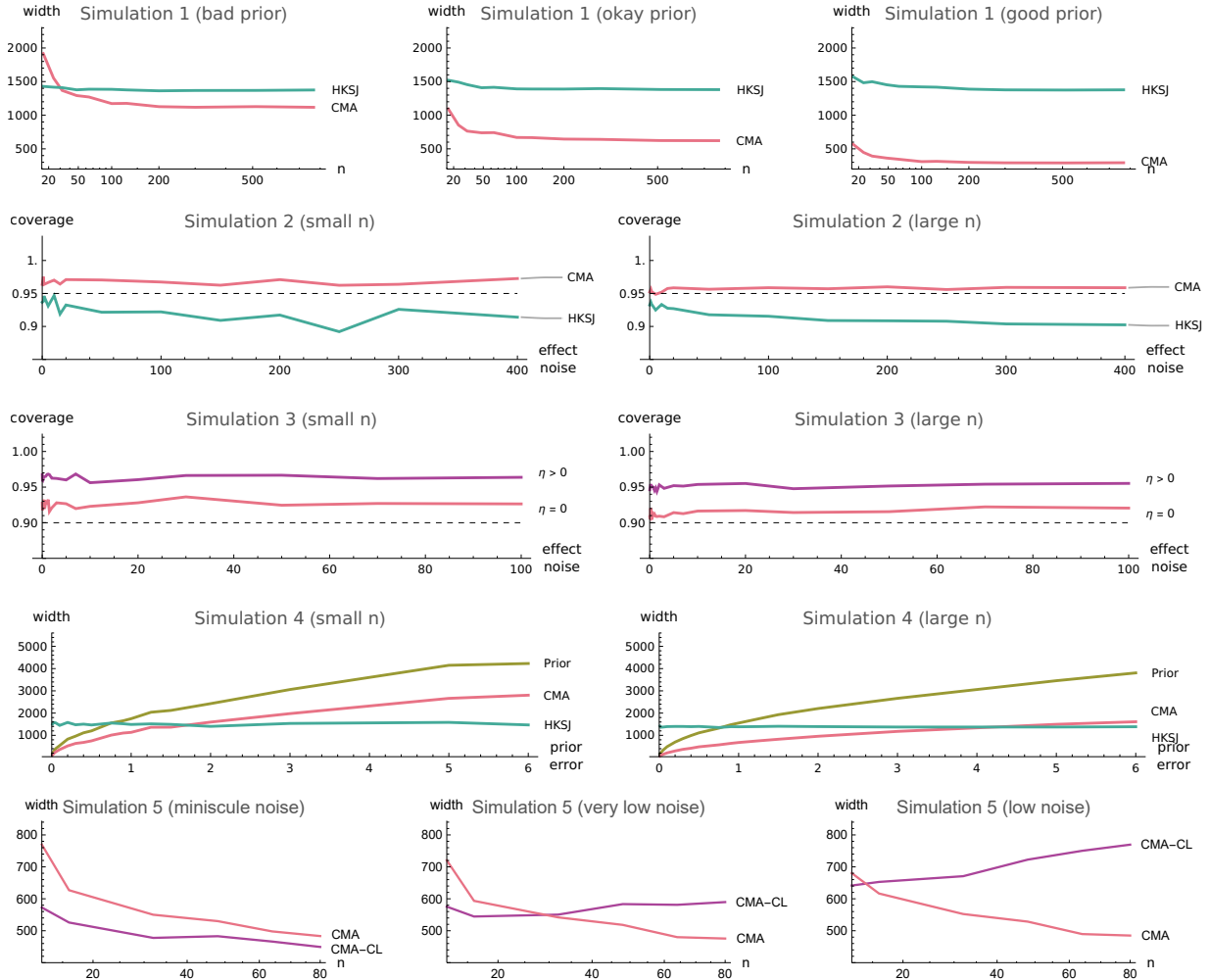

Figure 6: Results of all simulations on a single exemplar dataset. See Appendix A.1 for congruent results on the other datasets, as well as precise descriptions of the effect noise and prior error parameters. Overall, conformal meta-analysis can deliver much tighter intervals than traditional methods (Simulation 1), even though traditional methods have weak coverage guarantees (Simulation 2), whereas our algorithms, or their analyses, have (overly) strong guarantees (Simulation 3). Algorithm 5, not just a good prior, is essential to this performance (Simulation 4). Algorithm 3 can be better when there is very little noise (Simulation 5).

With the higher setting of $\eta$, over-coverage is consistently demonstrated. This suggests that the analysis of Section 5 can be improved, at least in some settings.

**Simulation 4**: Our approach assumes that, in many fields, it should be possible to develop good priors from large volumes of untrusted data. However, if these priors are indeed very accurate, it is unclear whether using KRR (upon just $n$ trials) is worth the complexity, and possible statistical overhead, over just using the prior as a fixed predictor. (This is conceptually equivalent to using an extremely large ridge parameter $\lambda$, or performing split conformal using all the training data for calibration.) This simulation indicates there is no such overhead: our fully-conformal intervals are strictly superior to those derived from a fixed prior. Thus, unless assumptions stronger than exchangeability are used to derive prediction intervals, learning is superior to mere validation. Note that, when prior error is high, HKSJ becomes superior.

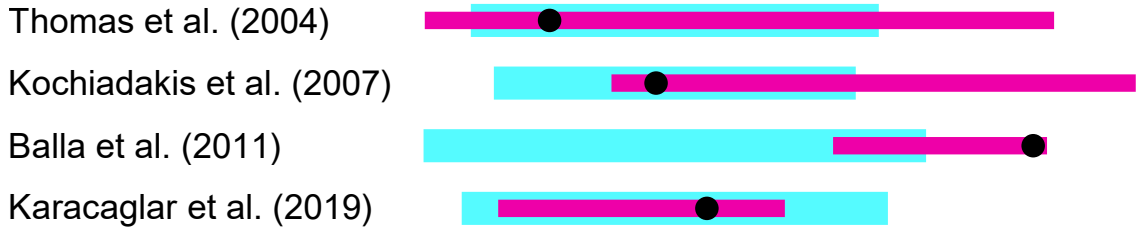

Figure 7: Prediction intervals for new observed effects $y$ (black dots) produced by traditional meta-analysis (light blue) and conformal meta-analysis (magenta, thin). On average, they are comparable in width (1.34 and 1.31, respectively). Conformal meta-analysis manages to cover the discrepant trial of Balla et al. [2011]. Note that the prior for conformal meta-analysis was produced **post-hoc**, having already seen the analysis of Letelier et al. [2003] and the results therein. Thus, these intervals should not be interpreted as quantitative evidence, but merely as qualitative illustrations of the behavior of conformal meta-analysis.

**Simulation 5**: Algorithm 5 is recommended for practical use, which is why it is used throughout the rest of the experiments. Nonetheless, Algorithm 3 was presented as a superior alternative in the (somewhat theoretical) scenario when the effects are "clean" (i.e. the trials are very large). This simulation experimentally confirms that Algorithm 3 can be superior in this setting. When there is essentially no effect noise, Algorithm 3 (labeled as CMA-CL) obtains lower width. However, its intervals become substantially worse as effect noise or $n$ increase.

## 7 Case Study: Amiodarone

We revisit the systematic review of Letelier et al. [2003], which assessed the effectiveness of amiodarone for atrial fibrillation (AF) patients. Its outcome measure is the relative risk of normal sinus rhythm; that is, the probability of restoring normal rhythm when administered amiodarone, divided by the probability of restoration with placebo. The review involved $n = 21$ trials, which we use as training data. For test data, we identify 4 trials that were published after the review, but would have met its inclusion criteria [Thomas et al., 2004, Kochiadakis et al., 2007, Balla et al., 2011, Karaçağlar et al., 2019]. Per the Predicting Trials task, we compare traditional meta-analysis (the Bayesian algorithm of Proposition 3, described in Section 2.6) with conformal meta-analysis (Algorithm 4, with $\eta = 1$).

Our goal is not to make any scientific claims about amiodarone, nor to reassess its evidence base; that would require following a formal, preregistered protocol. Though we temper our quantitative findings (depicted in Figure 7), we find them qualitatively interesting. Conformal meta-analysis manages to correctly predict all 4 trials, whereas traditional meta-analysis suffers a misprediction. This is not statistically convincing, but it aligns with the fact that conformal meta-analysis has a rigorous coverage guarantee, whereas traditional algorithms do not. (See Section 2.6 for more details). It is interesting to observe that not all of the conformal intervals overlap; by contrast, traditional intervals all inherently overlap. This suggests users of conformal meta-analysis could enjoy predictions that are meaningfully responsive to the details of their proposed treatment, perhaps distinguishing between effective and ineffective ones.

Appendix A.2 describes how we conducted the conformal meta-analysis. We highlight some ways it differed from the usual process. The first change is training a prior on helpful data that would otherwise be ignored. We identify 8 trials that did not meet the inclusion criteria, since they were not placebo controlled. To generate pseudo-effects for these trials, we need to understand the placebo effect. This leads to the second major change, which is holistically including the perspectives of practitioners. The critique of Slavik and Zed [2004], written by two doctors of pharmacy, gave estimates for the placebo effect on sinus rhythm (i.e. spontaneous conversion) in different circumstances. We use these estimates to generate the pseudo-effects. Finally, arguably the biggest change involves LLMs. In order to extract features from trials, we give their published PDFs to LLMs (specifically, GPT-4 and Claude) along with a prompt including example output. Next, parsing code (also written by LLMs) converts the textual features to numerical $(x, y, v)$. Thus, LLMs can be used to aid meta-analysis, much as meta-analysis serves as a question-answering system. This experience, and the results of the paper overall, reflect positively on the following dilemma: *can language models be used to rigorously answer scientific questions?*

# 8 Conclusion

This paper resolves two longstanding problems with meta-analysis: (1) how to derive rigorous conclusions from the untrusted, lower levels of the evidence hierarchy, and (2) how to manage pervasive heterogeneity when pooling together randomized controlled trials. It develops fundamentally more powerful meta-analysis algorithms based upon novel insights about full conformal prediction in the presence of noise. We believe that our approach has the potential to increase the accuracy of evidence synthesis, enhance the development of clinical practice guidelines, and ultimately improve patient outcomes. Furthermore, we believe our approach could make evidence-based medicine more harmonious and inclusive, by deprecating potentially divisive, controversial data-exclusion practices which are presently rationalized as necessary for statistical rigor.

However, further work must be conducted before this potential can be realized. Conformal meta-analysis depends upon priors derived from untrusted data, but such priors have yet to be developed. Conformal meta-analysis should ideally be paired with ongoing efforts to develop foundation models from large healthcare databases. From a technical perspective, this paper is meant to initiate the study of conformal meta-analysis, not to definitively solve it. The simulations indicate that the intervals of Algorithm 3 and Algorithm 5 are unnecessarily loose; a more fine-grained, thorough statistical examination of meta-analysis is warranted. Also, it should be possible to obtain comparably tight intervals under weaker assumptions. Specifically, the assumption of exchangeability could be relaxed, as in previous work on conformal prediction [Barber et al., 2023, Gibbs and Candès, 2024].

# A Appendix

## A.1 Simulation Details and Full Results

The simulations were performed using three partially-synthetic biomedical datasets from the Penn Machine Learning Benchmark [Olson et al., 2017]: 1196_BNG_pharynx, 1201_BNG_breastTumor, and 1193_BNG_lowbwt. We randomly subsample training data $(X, U)$ as well as test data $(x, u)$. The kernel matrix $K$ is generated using either the Gaussian or Laplace kernel as $\kappa$. For consistency across datasets having different scales, a parameter effect noise $> 0$ is introduced, and the distribution of $V$ is constructed to satisfy effect noise $= \mathbb{E}(V_i)^2/\mathbb{E}|U_i|$. Specifically $V_i \sim \mathrm{Exp}(1) \cdot \sqrt{\text{effect noise} \cdot \mathbb{E}|U_i|}$. Similarly, to produce prior means $M$ of varying quality, a parameter prior error $> 0$ is introduced, and the distribution of $M$ satisfies $\mathrm{MSE}(M, U) =$ prior error $\cdot \mathbb{V}(U)$. Furthermore, the difference between $M$ and $U$ should not be purely random — otherwise, using KRR to explain this difference would be hopeless. Instead, we generate a random offset function $\tilde{f}(x) = \sum_i g_i \kappa(\tilde{x}_i, x)$ for random held-out data $\tilde{x}_i$ and $g_i \sim N(0, 1)$. Since $\tilde{f}$ is an RKHS element generated from random data, there is some hope in approximating it using the training data. Letting $\tilde{F}$ be $\tilde{f}$ applied to the training features, we generate $M = p\tilde{F} + (1-p)U$ where $p = \sqrt{\text{prior error} \cdot \mathbb{V}(U)/\mathrm{MSE}(U, \tilde{F})}$.

All simulations are averaged over 32 random splits. Intervals are computed for between 256 and 768 test data in each run. Due to the efficiency of our proposed algorithms, all experiments are capable of running on a free Google Colab instance.

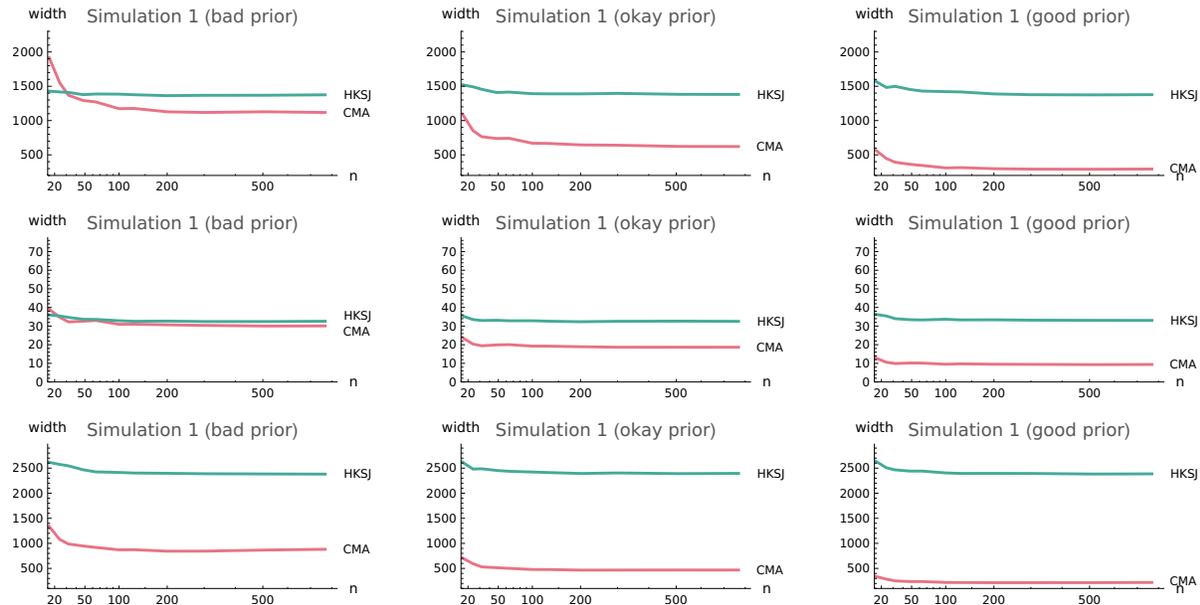

Simulation 1: Rows are different datasets; the different columns, from left to right, set prior error equal to 3.0, 0.9, and 0.2, respectively. $\alpha = 0.1$ and effect noise $= 0.5$ were used.

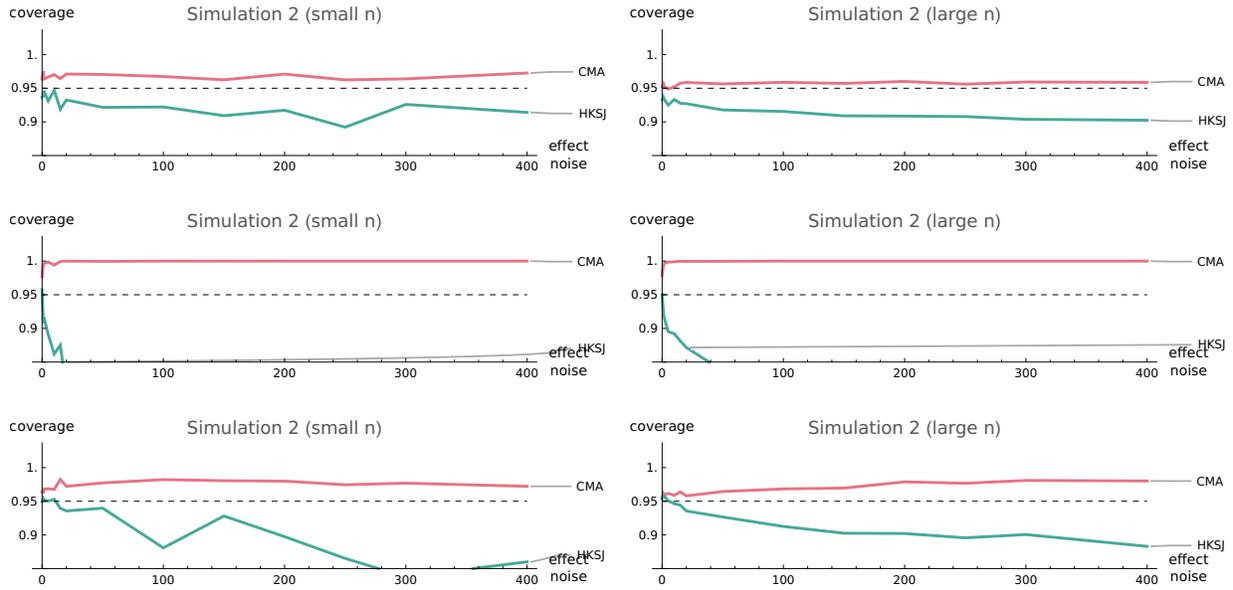

Simulation 2: Rows are different datasets. $n = 50$ and $n = 200$ are used in the left and right columns, respectively. prior error is set low to $0.2$.

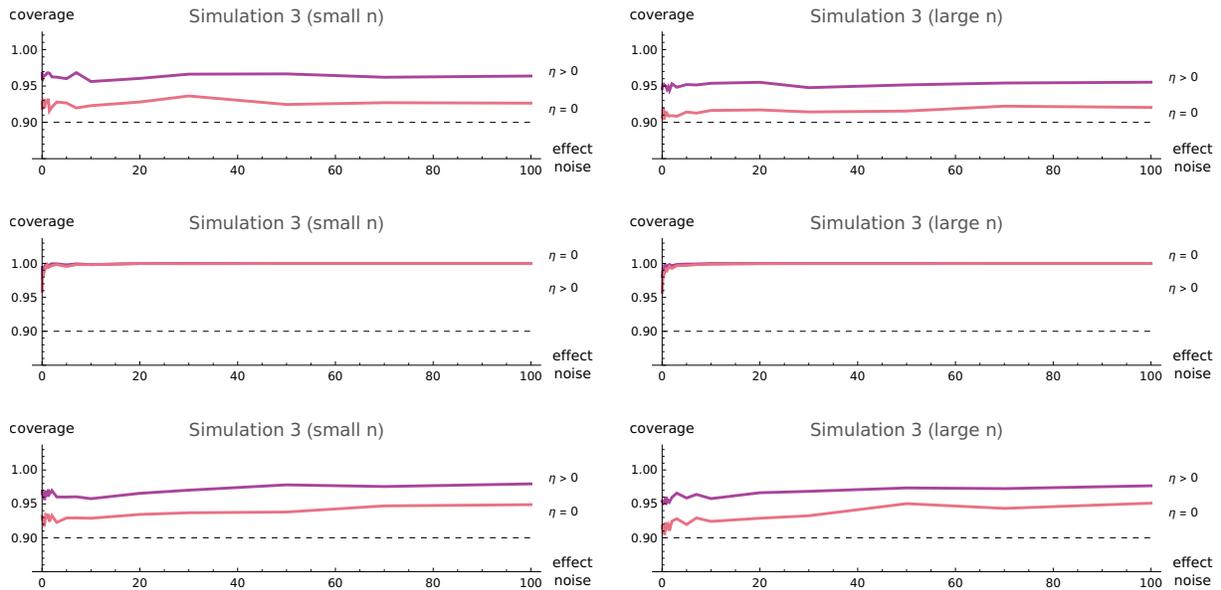

Simulation 3: Rows are different datasets; $n = 50$ and $n = 200$ are used in the left and right columns, respectively. $\alpha = 0.1$ and prior error $= 0.1$ were used.

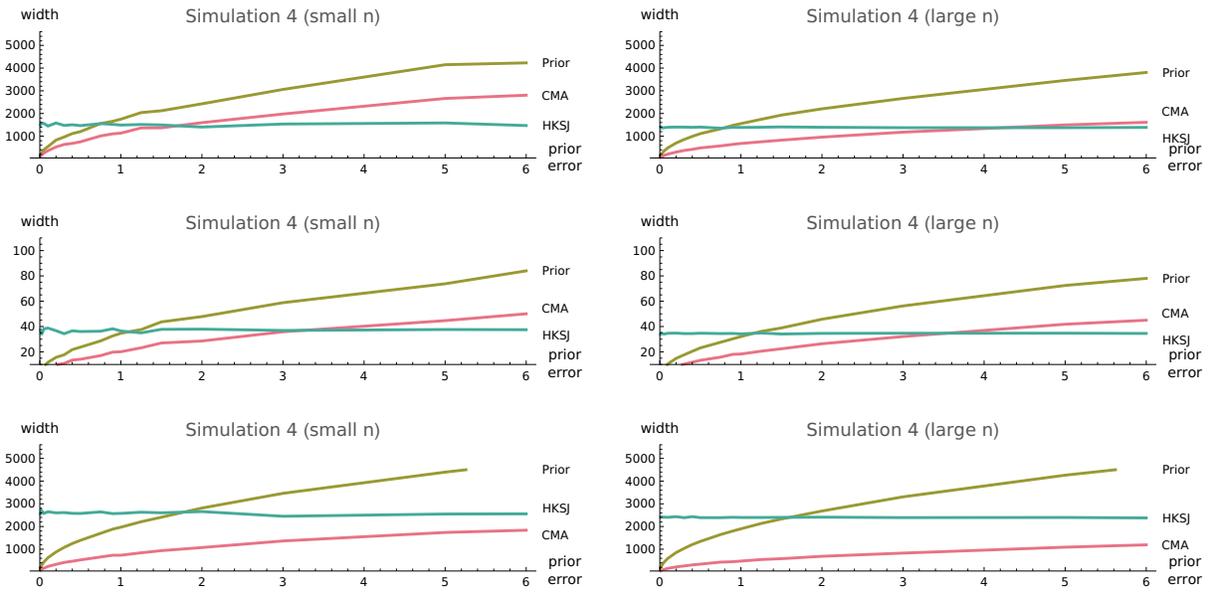

Simulation 4: Rows are different datasets; $n = 16$ and $n = 200$ are used in the left and right columns, respectively. A low effect noise $= 0.02$ was set, along with $\alpha = 0.1$.

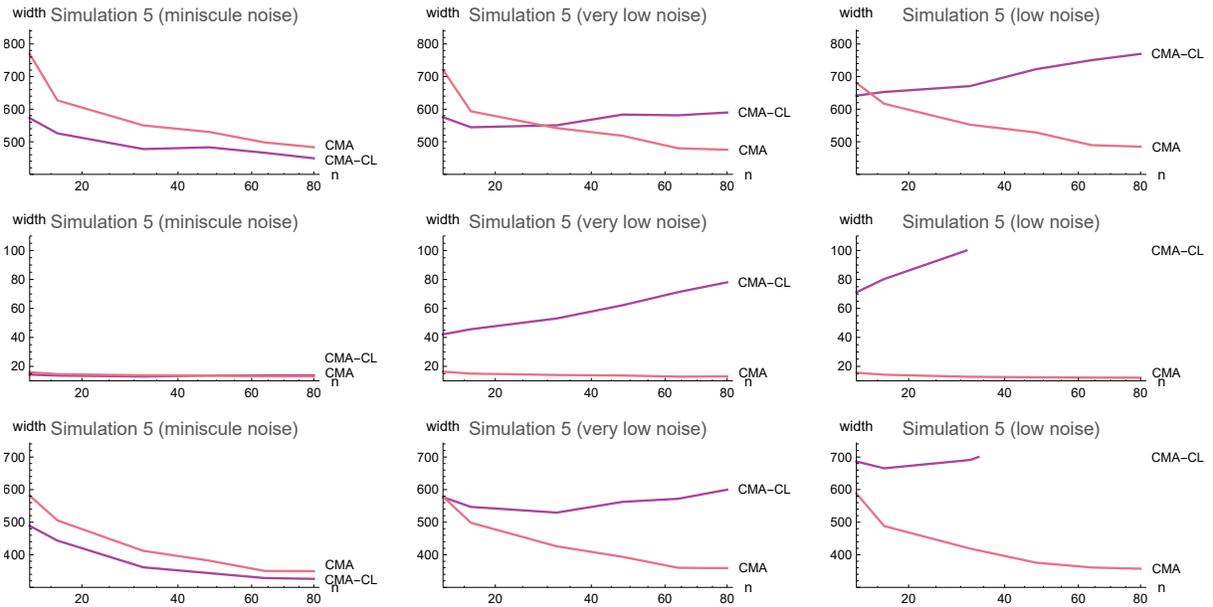

Simulation 5: Rows are different datasets; effect noise is set to $0.00000005$, $0.1$, and $2$ in the left, middle and right columns, respectively. A low confidence $\alpha = 0.25$ is set.

## A.2 Case Study Details

We follow the meta-analysis process illustrated in Figure 2. First, we determine the domain $\mathcal{X}$ of $x$. Helpfully, Letelier et al. [2003] identified 10 potentially-relevant features, such as mean age, mean AF duration, and amiodarone therapy protocol (e.g. "IV, 5 mg/kg in 30 min + 10 mg/kg in 20 h" or "Oral, 600 mg/d for 3 wk"). In order to extract these features from the trial, we give their published PDFs to a publicly-available language model, along with a prompt including example output. This extraction is fairly reliable, echoing the experience of Yun et al. [2024]. Next, parsing code (also written by the language model) converts the extracted textual features to numerical vectors $x$. As exemplified in Figure 10, this parsing can be tedious and error-prone, even with a state-of-the-art LLM. Our final predictions end up relying on just three features: total amiodarone dosage in the first 24 hours, whether mean AF duration was above or below 48 hours, and the number of patients (which is a sensible feature when predicting trials rather than effects).

In lieu of a powerful pretrained foundation model, we base $\mu$ and $\kappa$ on the critique of Slavik and Zed [2004]. They describe how multiple sources of heterogeneity, such as dosage, could impact the effect of amiodarone. Most importantly, amiodarone has a relatively slow course of action, whereas patients with recent-onset AF (usually defined as an AF duration of less than 48 hours) have a high chance of spontaneously reverting to normal sinus rhythm. (Letelier et al. [2003] also noted this pattern). With recent-onset AF, median spontaneous conversion rates are "11% at 2 hours after admission, 18% at 3 hours, 25% at 4 hours, 31% at 6 hours, 39% at 8 hours, 38% at 12 hours, 58% at 24 hours, and 67% at 48 hours.". This compares to only 0–8% within the first 72 hours for patients with persistent AF. We identify 8 further trials which compared amiodarone to an active comparison. We compute pseudo-effects (as relative risk) by taking the ratio of the observed probability of conversion under amiodarone, over the aforementioned estimated probability of spontaneous conversion over time. Such indirect comparison is reminiscent of how network meta-analysis works [Cipriani et al., 2013]. We trained a ReLU deep network upon the 3 relevant features in these synthetically-labeled data.

```
Can you extract the following features from the attached PDF paper? I gave example values, from another paper, which
should be replaced with the actual values in this paper. The only relevant outcome is conversion to normal sinus
rhythm. Also, create a new key like "Results": [a, b, c, d] where a is the number of amiodarone patients converted to
 sinus rhythm, b is the total number of amiodarone patients, c is the number of comparison patients converted to
sinus rhythm, and d is the total number of comparison patients. Answer as JSON.

{"Name": "Villani et al.11 (Italy) 2000", "Features": { "Amiodarone Therapy Protocol": "Oral, 400 mg/d for 1 mo", "
Comparison Treatment": "Oral digoxin, 0.25 mg/d or oral diltiazem hydrochloride 180- 360 mg/d for 1 mo", "Time to
Outcome Measure": "1 mo", "Number of Amiodarone Patients": "44", "Number of Control Patients": "30", "Fraction with
CV Disease": "47", "Mean Left Atrium Size, mm": "50", "Mean AF Duration": "17 wk", "Mean Age": "58", "Fraction Male":
 "67", "Adequate Concealment of Treatment": "No", "Follow-up Fraction": "100", "Masked Patients": "Yes", "Masked
Caregiver": "no", "Masked Assessor": "no" }}
```

Figure 8: Prompt used to extract relevant data from trial PDFs.

```
In the attached JSON list, each element represents a study described by the "Features" attribute. Convert these
features to real numbers so they can be provided to a learning algorithm.

* amiodarone treatment should be the total dosage, in milligrams, which is given over the first 24 hours. If the
dosage is specified per kg bodyweight, then take into account the average bodyweight of the patients.

* comparison treatment should be converted to [0,1], where 0 denotes placebo and 1 an intensive, high dose comparison
 regimen.

* if the fraction of male patients is unknown, just assume it is 0.5.

* fraction with CV disease and followup fraction were reported as integers, so for example 78 should be converted to
0.78.

* number of control and amiodarone patients should be just copied over as integers

* mean AF duration and time to outcome measure should be converted to -1 for <= 48 hours and 1 for > 48 hours

* mean left atrium size and mean age should be rescaled to \[-1,1\] where 0 is the average of the feature, -1 is the
minimum, and 1 is the maximum

* the boolean features should be rescaled to \[-1, 1\], where -1 means false, 1 means true, and 0 means not present
or not confident.

* include the same keys for all the studies, using the original key names.

Answer as JSON; no further explanation is necessary.
```

Figure 9: Prompt used to convert extracted data to numerical features.

```python
1   def parse_dosing_protocol(protocol):
2       if protocol is None or protocol.lower() == 'not specified':
3           return 0
4
5       weight = 70  # Average body weight in kg
6       total_mg = 0  # Initialize total milligrams
7
8       # Normalize and break down the protocol into components
9       protocol = protocol.lower().replace('over', 'in').replace('plus', ',')
10      phases = protocol.split('+')
11
12      for phase in phases:
13          parts = phase.split(',')
14          for part in parts:
15              part = part.strip()
16              tokens = part.split()
17              dose = 0
18              rate_based = False
19              duration = 24  # Default duration is 24 hours unless specified
20
21              # Parse the dose and units
22              for i, token in enumerate(tokens):
23                  try:
24                      # Attempt to convert token to float to find numeric values
25                      potential_dose = float(token)
26
27                      # Check for units immediately following the numeric value
28                      if i + 1 < len(tokens):
29                          unit = tokens[i + 1]
30                          if 'g' in unit and 'mg' not in unit:
31                              potential_dose *= 1000  # Convert grams to milligrams
32                          elif 'mg/kg' in unit:
33                              potential_dose *= weight  # Convert to total mg for given
                                  ↪   weight
34
35                          # Determine if the dose is time-bound
36                          if 'hour' in unit or 'h' in unit or 'min' in unit:
37                              rate_based = True  # The dose is a rate per time
38                              duration = extract_duration(part)
39                              if 'min' in unit:
40                                  duration /= 60  # Convert minutes to hours
41                      dose = potential_dose
42                      break
43                  except ValueError:
44                      continue  # Not a number, move to next token
45
46              # Apply the dose calculation based on the duration and whether it's
                  ↪   rate-based
47              if rate_based:
48                  total_mg += min(duration, 24) * dose  # Apply the rate up to 24 hours
49              elif 'day' in part:
50                  if 'first' in part or '1 day' in part or '1 week' in part:
51                      total_mg += dose  # Apply if it specifies the first day or week
52                  else:
53                      total_mg += dose  # Single dose or calculated for the duration
54
55      return total_mg
```

Figure 10: Python code generated by GPT-4 to parse and convert amiodarone therapy protocols. Generating this code required multiple rounds of interaction with the language model. This code still has mild bugs, which are left untouched to accurately convey contemporary expectations about in-context parsing.